\documentclass[letterpaper]{article} % DO NOT CHANGE THIS
\usepackage{aaai2026}  % DO NOT CHANGE THIS
\usepackage{times}  % DO NOT CHANGE THIS
\usepackage{helvet}  % DO NOT CHANGE THIS
\usepackage{courier}  % DO NOT CHANGE THIS
\usepackage[hyphens]{url}  % DO NOT CHANGE THIS
\usepackage{graphicx} % DO NOT CHANGE THIS
\usepackage{natbib}  % DO NOT CHANGE THIS AND DO NOT ADD ANY OPTIONS TO IT
\usepackage{caption} % DO NOT CHANGE THIS AND DO NOT ADD ANY OPTIONS TO IT
\usepackage[utf8]{inputenc} % allow utf-8 input
\usepackage[T1]{fontenc}    % use 8-bit T1 fonts
\usepackage{url}            % simple URL typesetting
\usepackage{booktabs}       % professional-quality tables
\usepackage{amsfonts}       % blackboard math symbols
\usepackage{nicefrac}       % compact symbols for 1/2, etc.
\usepackage{microtype}      % microtypography
\usepackage{caption}
\usepackage{xcolor}         % colors
\usepackage{url}
\usepackage[utf8]{inputenc} % allow utf-8 input
\usepackage[T1]{fontenc}    % use 8-bit T1 fonts
\usepackage{url}  
\usepackage{amsmath}
\usepackage{floatrow}
\usepackage{caption}
\usepackage{subcaption}
\usepackage{enumitem}
\usepackage{natbib}
\usepackage{graphicx}% simple URL typesetting
\usepackage{booktabs}       % professional-quality tables
\usepackage{amsfonts}       % blackboard math symbols
\usepackage{nicefrac}       % compact symbols for 1/2, etc.
\usepackage{microtype}      % microtypography
\usepackage{xcolor}
\usepackage{enumitem}% colors
\usepackage{multirow}
\usepackage{graphicx}
\usepackage{booktabs}
\usepackage{multirow}
\usepackage{bbding}
\usepackage{amsmath}
\usepackage{tabularx}
\usepackage{longtable} % For tables spanning multiple pages
\usepackage{algorithm}
\usepackage{algorithmic}

\usepackage{newfloat}
\usepackage{listings}
\DeclareCaptionStyle{ruled}{labelfont=normalfont,labelsep=colon,strut=off} % DO NOT CHANGE THIS
\floatstyle{ruled}
\newfloat{listing}{tb}{lst}{}
\floatname{listing}{Listing}
\title{Harnessing Diffusion-Generated Synthetic Images for Fair Image Classification}
\author{
    Abhipsa Basu\textsuperscript{\rm 1},
    Aviral Gupta\textsuperscript{\rm 2},
    Abhijnya Bhat\textsuperscript{\rm 3},
    R. Venkatesh Babu\textsuperscript{\rm 1},
}
\affiliations{
    \textsuperscript{\rm 1}Indian Institute of Science, Bangalore
    \textsuperscript{\rm 2}BITS Pilani
    \textsuperscript{\rm 3}Stanford University
}

\usepackage{bibentry}
\begin{document}

\maketitle
\begin{abstract}
Image classification systems often inherit biases from uneven group representation in training data. For example, in face datasets for hair color classification, blond hair may be disproportionately associated with females, reinforcing stereotypes. A recent approach leverages the Stable Diffusion model to generate balanced training data, but these models often struggle to preserve the original data distribution. In this work, we explore multiple diffusion-finetuning techniques, e.g., LoRA and DreamBooth, to generate images that more accurately represent each training group by learning directly from their samples. Additionally, in order to prevent a single DreamBooth model from being overwhelmed by excessive intra-group variations, we explore a technique of clustering images within each group and train a DreamBooth model per cluster. These models are then used to generate group-balanced data for pretraining, followed by fine-tuning on real data. Experiments on multiple benchmarks demonstrate that the studied finetuning approaches outperform vanilla Stable Diffusion on average and achieve results comparable to SOTA debiasing techniques like Group-DRO, while surpassing them as the dataset bias severity increases.
\end{abstract}

\begin{links}
    \link{Code}{https://github.com/abhipsabasu/harnessing_diff_models}
    % \link{Datasets}{https://aaai.org/example/datasets}
    % \link{Extended version}{https://aaai.org/example/extended-version}
\end{links}

% \vspace{-0.5cm}
\section{Introduction}
\label{sec:intro}

Image classification models often exhibit harmful biases, posing significant risks for real-world deployment~\cite{wang2021gender, zhao2017men, metaxa2021image}. These biases arise from imbalances in training data; e.g., in CelebA~\cite{celeba}, blond female faces considerably outnumber blond males, leading to misclassification of the latter. While numerous debiasing techniques have been proposed~\cite{sagawa2019distributionally, kirichenko2022last, nam2020learning}, mitigating bias becomes increasingly tough when dataset imbalances become severe. With the recent breakthroughs in image generation using models like Stable Diffusion~\cite{ramesh2022hierarchical}, we pose a critical question: \textit{Can we harness the generative power of such models to create images that facilitate the training of fair classification systems, even in presence of extreme dataset bias?}

A recent work FFR~\citep{qraitem2023fake} leverages images generated by Stable Diffusion (SD) to train fair classification systems. However, due to the stochastic nature of diffusion models~\cite{shin2023fill}, SD-generated images often diverge from the original data distribution and may not follow prompt instructions accurately. Consider the Waterbirds dataset~\cite{sagawa2019distributionally}, which contains `waterbird' and `landbird' classes. Spurious correlations arise as classification models associate water backgrounds with waterbirds and land backgrounds with landbirds, thus relying on background rather than the bird features for predictions. Attempts to prompt SD to generate `waterbird on land' images often produce water backgrounds, even when explicitly instructed otherwise (Appendix Fig.~\ref{fig:wb}). FFR attempts to mitigate this by using highly specific prompts, such as `\texttt{photo of a flamingo on pavement}', relying on exact bird names and background details. However, without precise domain knowledge, such prompts may yield irrelevant or out-of-distribution images. In this work, we explore fine-tuning generative models directly on the dataset, enabling them to better capture the data distribution and significantly improve classification fairness.

\begin{figure*}[t!]
    \centering
    \includegraphics[width=0.85\textwidth, trim=0 230 0 105, clip]{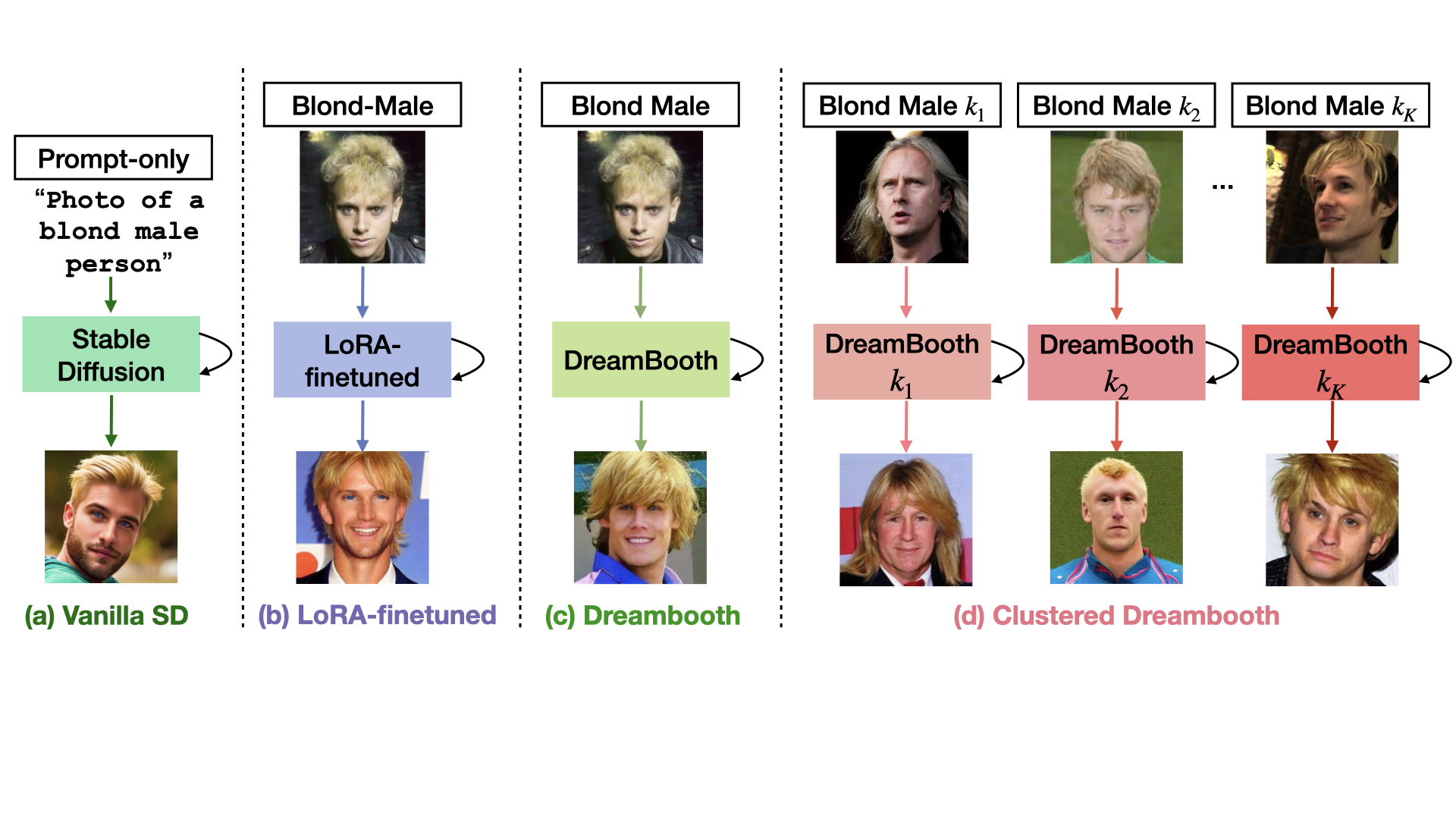}
    \caption{\textbf{Generation Approaches.} In this paper, we investigate four image generation pipelines for training fairer classifiers: a) \textit{Vanilla SD}, which only accepts prompts, b) \textit{LoRA-finetuning}, which finetunes the diffusion model on the images of a group, c) \textit{Dreambooth}, which finetunes the diffusion model and represents the images of a group using a special token `\texttt{V}', d) \textit{Clustered Dreambooth}, which finetunes a diffusion model on different clusters $k_i$ present in the training group, representing each of them through a special token `\texttt{V}'.}
    \label{fig:teaser}
\end{figure*}

To generate images closely aligned with the training set, we fine-tune Stable Diffusion using LoRA~\cite{lora} on each dataset group (e.g., Non-Blond Females, Blond Males in CelebA~\cite{celeba}). Prior work on synthetic data augmentation~\cite{shin2023fill} trains personalized diffusion models on individual classes to produce class-consistent, in-distribution images. Building on this, we further explore DreamBooth~\cite{ruiz2022dreambooth}, which introduces a special token (`\texttt{[V]}') to represent the subject, improving resemblance to real samples. While such models typically target visually similar objects (e.g., a specific dog breed), our training groups contain diverse images sharing only a common attribute (e.g., hair color). To address this intra-group variation, we explore a simple extension—Clustered DreamBooth—which clusters each group into visually similar subsets and fine-tunes a separate DreamBooth model per cluster. These different generation methods are illustrated in Fig.~\ref{fig:teaser}. Using each strategy, we generate equal number of images per group. 

After generating group-wise images, the key question becomes: \textit{how should we use them for training?} FFR addresses this by first training a classification model on the synthetic data, followed by finetuning on the original dataset. However, since the original data is biased, this step risks reintroducing bias—unless extensive hyperparameter tuning is performed, as in FFR. To avoid this, we propose to finetune only the classifier's softmax layer with the real data~\cite{basu2024mitigating}. Both the training and finetuning stages are optimized using a weighted combination of Cross-Entropy and Supervised Contrastive Loss~\cite{khosla2020supervised}, which enhances feature representations by promoting better class separation.

We evaluate our proposed pipeline on three fairness benchmarks—Waterbirds~\cite{sagawa2019distributionally}, CelebA~\cite{celeba}, and UTKFace~\cite{zhang2017age}. The training on vanilla SD generated images followed by final layer retraining improves the accuracy of the classification systems over FFR indicating the utility of our finetuning approach. The different finetuning strategies discussed in the paper also offer added advantage for all the datasets, performing comparably to traditional debiasing techniques like GroupDRO~\cite{sagawa2019distributionally} and SELF~\cite{labonte2023towards}. Notably, as dataset biases intensify, such methods deteriorate, while our generation-based approaches remain robust, demonstrating their effectiveness in highly imbalanced settings. We summarize the key contributions below:
% \vspace{-0.3cm}
\begin{itemize}[leftmargin=*, itemsep=0pt]
    \item We explore the utility of diffusion models and corresponding finetuning mechanisms like LoRA and Dreambooth, which learn to generate images from those in the biased training groups, to develop fair classifiers. We also explore Clustered Dreambooth, that first clusters group images and then trains distinct Dreambooth models on each cluster to better capture intra-group variations. 
    \item We propose a two-stage pipeline for leveraging synthetic data for building fair classification systems. We generate equal number of images per group using our approaches and train a classification model on them. To further enhance performance, we finetune only the softmax layer of the model with the original data.
    \item Through extensive experiments on multiple benchmarks, we show that all diffusion-based methods enhance the training of fair classifiers more effectively than FFR. Notably, they surpass debiasing methods like Group-DRO by a large margin, when the dataset bias ratios become severe.
\end{itemize}
% \vspace{-0.2cm}
\section{Related Work}
\label{sec:related}
\noindent \textbf{Bias Mitigation} has been widely studied, with approaches falling into two categories: \textit{known} and \textit{unknown biases}. For \textit{known} biases, the spurious attribute is known apriori~\cite{kim2019learning, li2019repair, sagawa2019distributionally, arjovsky2019invariant, teney2021unshuffling, tartaglione2021end, wang2020towards, Wang_2022_CVPR, basu2023rmlvqa}. In GroupDRO~\cite{sagawa2019distributionally}, the worst-group training loss is optimized. Last Layer Retraining~\cite{kirichenko2022last} shows that pretraining the model on the biased dataset and then only retraining the final classification layer with a group-balanced validation set can help debias the model. Semi-supervised approaches assume bias annotations only for a few samples~\cite{nam2020learning, jung2022learning}. For \textit{unknown} biases, where bias attributes and labels are not available~\cite{creager2021environment, lee2021learning, Li_2022_ECCV, lahoti2020fairness, ahn2023mitigating, liu2021just, huang2020self, hong2021unbiased, darlow2020latent, basu2024mitigating}, some approaches use dual-branch networks—one amplifying bias, the other mitigating it~\cite{nam2020learning, lee2021learning, liu2022avoiding}. Contrastive-based methods~\cite{zhang2022correct, zhang2022contrastive} refine feature representations by clustering same-class samples, identifying pseudo bias labels via model misclassifications. Such biases are also exhibited by diffusion models~\cite{basu2023inspecting, parihar2024balancing}.

\noindent \textbf{Data Augmentation using Generative Models.} Many recent works utilize generative models for data augmentation~\cite{trabucco2023effective, azizi2023synthetic, du2024dream, zheng2023toward, mariani2018bagan}. The trend began with GANs~\cite{goodfellow2020generative}. BAGAN~\cite{mariani2018bagan} was used to augment class-imbalanced datasets to enhance minority class performance. With diffusion models, DA-Fusion~\cite{trabucco2023effective} employs Textual Inversion~\cite{gal2022image} to generate augmentations for images of every class, and then during training, in each batch, retain every original image with probability $p$ and an augmented image with probability $(1-p)$. DiffuseMix~\cite{Islam_2024_CVPR} combines a partial natural image and its generated counterpart from the diffusion model, and thereafter combats adversarial attacks by blending a randomly selected structural pattern from a set of fractal images into the concatenated image to form the final augmented version for training.

\noindent \textbf{Generative Models for Debiasing} utilize generative models to debias classification systems~\cite{an20222, ramaswamy2021fair, qraitem2023fake, sharmanska2020contrastive}. GAN-based approaches~\cite{an20222, ramaswamy2021fair} train generative models on the training images to synthesize bias-conflicting samples that can augment the original data. The diffusion-based methods do not train the models from scratch, rather manipulate existing pretrained models to generate group-balanced images. FFR~\cite{qraitem2023fake} likewise generates group-balanced images from Stable Diffusion, trains the classification model on this synthetic data, before finetuning the latter with real data. Our work additionally explores the use of different finetuning mechanisms in text-to-image models to generate in-distribution images directly from the training image groups.
% \vspace{-0.2cm}
\section{Problem Statement and Methodology}
\label{sec:method}
% \vspace{-0.2cm}

\begin{figure*}[h!t]
    \centering
    \includegraphics[width=0.9\textwidth, trim=10 50 10 10, clip]{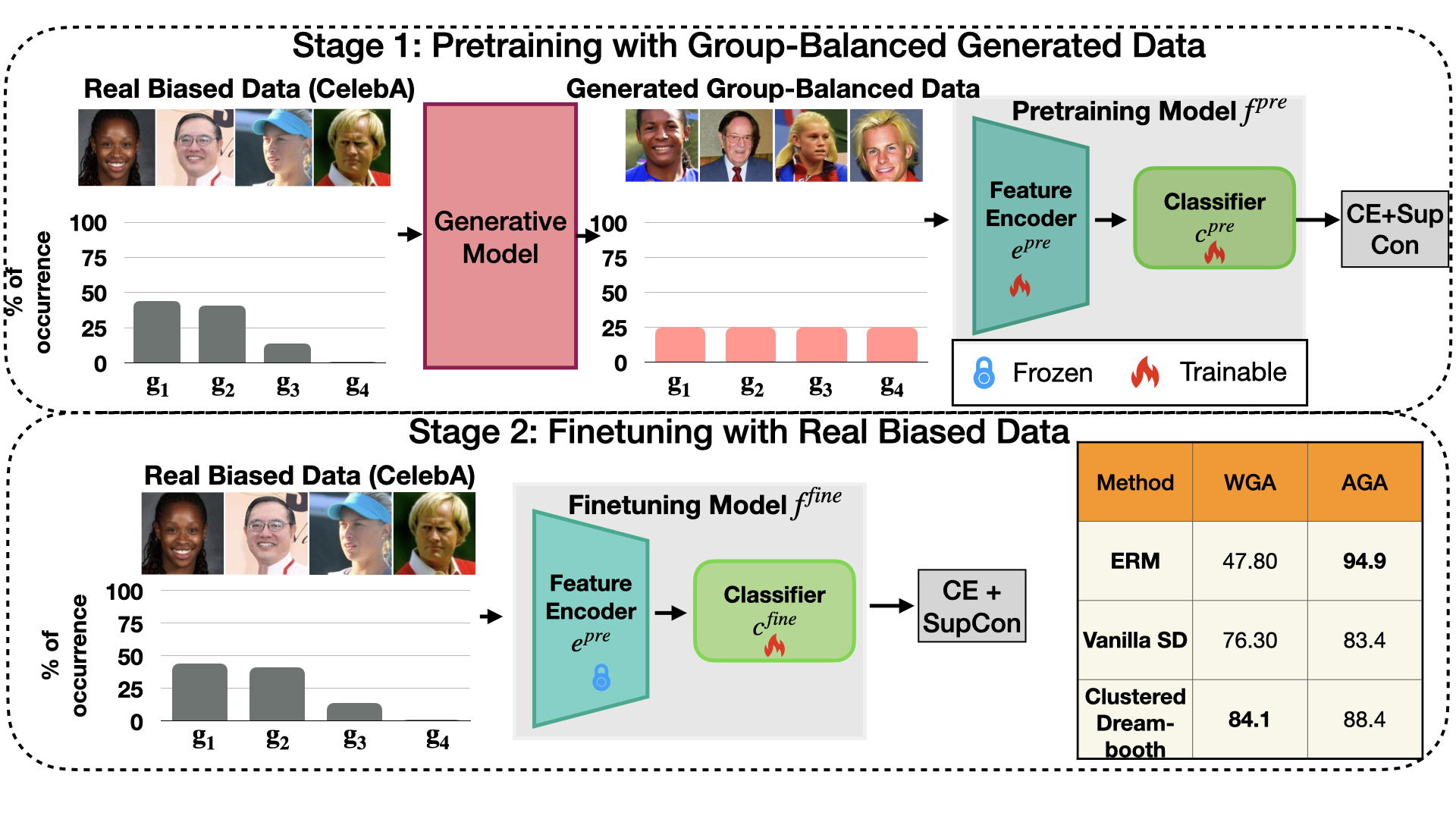}
    \caption{\textbf{Overview of the studied pipeline.} In Stage 1, we generate images uniformly from each group (e.g., non-blond female ($g_1$), non-blond male ($g_2$), blond female ($g_3$), blond male ($g_4$)) using the generatice approaches, and train a classification model $f^{\text{pre}}$ with CE and SupCon losses. In Stage 2, we finetune only the linear classifier on the original dataset.}
    \label{fig:main}
\end{figure*}

\noindent \textbf{Preliminaries}.
% \vspace{-0.2cm}
The motivation of this work is to train fairer image classification models using synthetic data. Let $\mathcal{X}$ be the set of real training images, where each $x_i \in \mathcal{X}$ is associated with a class label $y_i \in \mathcal{Y}$, a bias label $a_i \in \mathcal{A}$, and a group label $g_i \in \mathcal{G}$ where $g_i = (y_i, a_i)$. A mapping function $f:\mathcal{X} \rightarrow \mathcal{Y}$ is optimized by training a model in order to classify the images. Such a model has two parts: a) Feature encoder $e$, which is typically pretrained on a large dataset, and b) Classifier $c$ which is finetuned along with encoder $e$ to learn the class labels from the images. This model is traditionally trained using the Cross-Entropy (CE) loss (see Appendix~\ref{subsec: losses}).
% , which is defined below:
% \begin{equation}
% \label{eq:1}
%     CE = \frac{1}{|\mathcal{X}|}\sum_{i=1}^{|\mathcal{X}|}\sum_{j=1}^{|\mathcal{Y}|}-p_{ij}\log\hat{p}_{ij}
% \end{equation}
% where $[p_{i1}, p_{i2}, \cdots, p_{i|\mathcal{Y}|}]$ is the one-hot vector representation of $y_i$ and $[\hat{p}_{i1}, \hat{p}_{i2}, \cdots, \hat{p}_{i|\mathcal{Y}|}]$ is the corresponding softmax vector, obtained from the model $f$.
The model becomes biased when group frequencies in the training data are imbalanced, resulting in superior performance on some groups in the test data and inferior performance on others.

% \vspace{-0.15cm}
\subsection{Generating Synthetic Images}
\label{subsec: gen_role}
With advancements in generative modeling~\cite{song2020denoising, ramesh2022hierarchical}, we explore their potential in training fair classifiers by generating images that reflect the training distribution and improve generalization to minority groups. Below, we outline our image generation methods and strategies for leveraging synthetic data to enhance fairness.

\noindent \textbf{Vanilla Stable Diffusion (SD).} 
We generate images from each group $g = (y,a)$ by specifying only $y$ and $a$ in the prompts. Such generations are independent of the training data, leading to domain mismatches, and may result in inaccurate generations if the model fails to follow the text prompts precisely (see \S~\ref{sec:intro}). 

\noindent \textbf{LoRA-based Finetuned Stable Diffusion.} 
We make the generative model aware of the training data by finetuning the SD model on each training group $g$ separately. Each model is trained on $l = \min\{|g| : g \in \mathcal{G}\}$ samples, selected randomly from each training group.
The images are generated using the model trained on $g$ by specifying $y$ and $a$ in the prompt.

\noindent \textbf{Dreambooth}.
To strengthen the resemblance between the training and generated images, we explore Dreambooth~\cite{ruiz2022dreambooth}, a text-to-image personalization model that learns to imitate an object or a concept (e.g., a specific dog) from a small set of images depicting that object. It finetunes a pretrained text-to-image model by learning a unique identifier (e.g., ``\texttt{[V]}'') such that on inference time, if the model is queried by that identifier (e.g., ``\texttt{photo of a [V] dog}''), it generates new images of the given object. Likewise, we sample $100$ images from each training group, and train a separate Dreambooth model $h$ on each group, where the prompt is of the form ``\texttt{photo of a [V] \textit{y}}''.
% It is to be noted that the groups in biased training datasets do not always contain a specific object, rather, it is a diverse set of images sharing a few common traits (e.g., hair color, gender). One can train a Dreambooth model $h$ on this entire group of images in order to generate new samples of the given group. 

\noindent \textbf{Clustered Dreambooth}. Dreambooth specializes in learning a concept from $3-5$ images. However, a training group like Blond Male consists of images of many individuals sharing a common trait, hair color. To prevent overwhelming a single Dreambooth model with multiple images of a training group, we explore a simple extension of first clustering the CLIP embeddings~\cite{pmlr-v139-radford21a} of the images within each group. Let $k^g_D$ denote the number of clusters, where $D$ and $g$ refer to the training dataset and a group in $D$ respectively. We train a pool of Dreambooth models $\mathcal{H}^g = \{h^g_1, h^g_2, h^g_3, \cdots, h^g_{k^g_D}\}$ on the obtained clusters. We implement Clustered-Dreambooth (i.e., the Dreambooth pool $\mathcal{H}^g$) using LoRA-based finetuning~\cite{hu2021lora}, which ensures lesser, feasible training time. Finally the trained models are utilized to generate images for each $g$. For simplicity, we assume equal $k^g_D$ for each group $g$, and denote the number of clusters as $k_D$ for the rest of the paper.

\vspace{-0.1cm}
\subsection{Stage 1: Training with the Generated Images}
\vspace{-0.1cm}
Once the generative models are trained with the individual data groups, we generate $M$ images from each group $g$ using Vanilla SD, LoRA-finetuned SD and Dreambooth. For Clustered Dreambooth, we generate $M_D^{cl}$ images from each cluster in a group belonging to dataset $D$, such that the total number of images generated from the group is $M^{cl}_D \times k_D = M$. However, inspite of training models on the dataset groups, all the generated images may not always follow the prompt or may not be of high quality. Thus a filtering step is required for selecting images appropriate for our training. This is done using a CLIP-based scoring mechanism.

\noindent \textbf{CLIP-based Filtering}. To find the most relevant images, we apply a CLIP score in two ways for each image $I$. 
\begin{enumerate}[leftmargin=*, itemsep=0pt]
    \item \textbf{CLIP-Label}$\mathbf{(I, p^c)}$ : We compute the image-text similarity of $I$ with a prompt $p^c$, of the format ``\texttt{Photo of a \{c\}}'', where $c$ is the class label.
    \item \textbf{CLIP-Centroid}$\mathbf{(I, \bar{z}^g)}$: To ensure that the chosen images most resemble the distribution in the given training group $g$, we compute the centroid of the CLIP embeddings of the images of the group, denoted by $\bar{z}^g = \frac{1}{M_g}\sum_{i=1}^{M_g}z_i^g$, where $M_g$ is the size of the group in the training set, and $z_i^g$ is the CLIP embedding of the $i^{th}$ image of $g$. We calculate the CLIP similarity between each generated image and its corresponding group centroid.  
\end{enumerate}

% \noindent $\text{CLIP-Label}(I, p^c)$ : We compute the image-text similarity of $I$ with a prompt $p^c$, of the format ``\texttt{Photo of a \{c\}}'', where $c$ is the class label.

% \noindent \textbf{CLIP-Centroid}$\mathbf{(I, \bar{z}^g)}$: To ensure that the chosen images most resemble the distribution in the given training group $g$, we compute the centroid of the CLIP embeddings of the images of the group, denoted by $\bar{z}^g = \frac{1}{M_g}\sum_{i=1}^{M_g}z_i^g$, where $M_g$ is the size of the group in the training set, and $z_i^g$ is the CLIP embedding of the $i^{th}$ image of $g$. We calculate the CLIP similarity between each generated image and the corresponding centroid.

\noindent The final scoring function becomes a combination of $\text{CLIP-Label}(I, p^c)$ and $\text{CLIP-Centroid}(I, \bar{z}^g)$:

\begin{align}
\label{eq:2}
\text{CLIP-Score}(I, p^c, \bar{z}^g) 
&= \alpha \cdot \text{CLIP-Label}(I, p^c) \notag \\
&\quad + (1 - \alpha) \cdot \text{CLIP-Centroid}(I, \bar{z}^g)
\vspace{-0.1cm}
\end{align}

\noindent where $\alpha$ is a hyperparameter. After selecting the top-ranked images from each group, a classification model $f^{\text{pre}}$ is trained on them to learn fair representations. An attentive reader may question the suitability of CLIP embeddings for filtering generated images. While this is indeed an important issue, we empirically validate the benefit of this filtering technique for our purposes by an experiment (Appendix Table~\ref{tab:alpha}).
% to denote that the synthetic images are generated from the generative model $h$.

\subsection{Stage 2: Finetuning with Original Data}
\label{subsec: finetune}
% \vspace{-0.1cm}
After pretraining the classification model on group-balanced synthetic data, we adapt it to the real data by finetuning it with the real images using the method described below.

\noindent \textbf{Last Layer Retraining with Real Data}. We finetune the trained model $f^{\text{pre}}$ on the entire real training dataset to help it adapt to real data. However, since the data is biased, finetuning the full network risks reintroducing the biases. To mitigate this, we freeze the feature encoder $e^{\text{pre}}$ and only finetune the linear classification layer $c^{\text{pre}}$. Additionally, to address any class imbalance, each finetuning batch samples classes uniformly. We refer to this method as LLR$_{\text{all}}$, and the finetuned model as $f^{\text{fine}}$. We differ here from FFR, where the entire model is finetuned, increasing the dependency on thorough hyperparameter tuning. Our two-stage approach is illustrated in Figure~\ref{fig:main} based on CelebA~\cite{celeba}.

We train both stages using a weighted sum of CE loss and Supervised Contrastive (SupCon) loss~\cite{khosla2020supervised} (see Appendix~\ref{subsec: losses}) to enhance class separation in both stages: $L = \beta. L_{\text{CE}} + (1-\beta) L_{\text{sup-con}}$, where $\beta = 0.5$ in all experiments.

\section{Experiments and Results}
\label{sec:experiments}
% \vspace{-0.1cm}

We present an overview of the datasets used for evaluation, followed by a comparative analysis of the diffusion model variants to determine their performances. 
% \vspace{-0.1cm}

\noindent \textbf{Datasets For Evaluation}:
Waterbirds~\cite{sagawa2019distributionally} is a dataset of bird images, labeled as waterbird if the bird is commonly seen with waterbodies on background, and landbird otherwise. The dataset, with $4,795$ samples, suffers from background bias: only a few waterbirds in the dataset have land on the background, and few landbirds have water. We also consider two real-world datasets. The CelebA dataset~\cite{celeba} consists of $202,599$ images of celebrities with annotations of $40$ binary attributes. We choose Blond Hair as the target attribute, which is known to suffer from gender biases~\cite{Li_2022_ECCV, seo2022unsupervised}. UTKFace~\cite{zhang2017age} is a dataset of human faces, having $20,000$ images with annotations of age, gender, and ethnicity. We use gender and age as the target and bias attributes respectively -- where female adults dominate female children, and male children dominate male adults. Further details on this data split is provided in Appendix~\ref{subsec:utk}. For each dataset, we report the worst (WGA) and average group (AGA) accuracies, following previous works~\cite{sagawa2019distributionally, labonte2023towards}.

% \subsection{Competing Generation Baselines}
% To evaluate the effectiveness of DreamDebias, we compare it against existing image-generation baselines (described below), including single Dreambooth~\cite{ruiz2022dreambooth} per group (DB).
\begin{figure}
    \centering
    \includegraphics[width=0.8\linewidth, trim=400 360 230 100, clip]{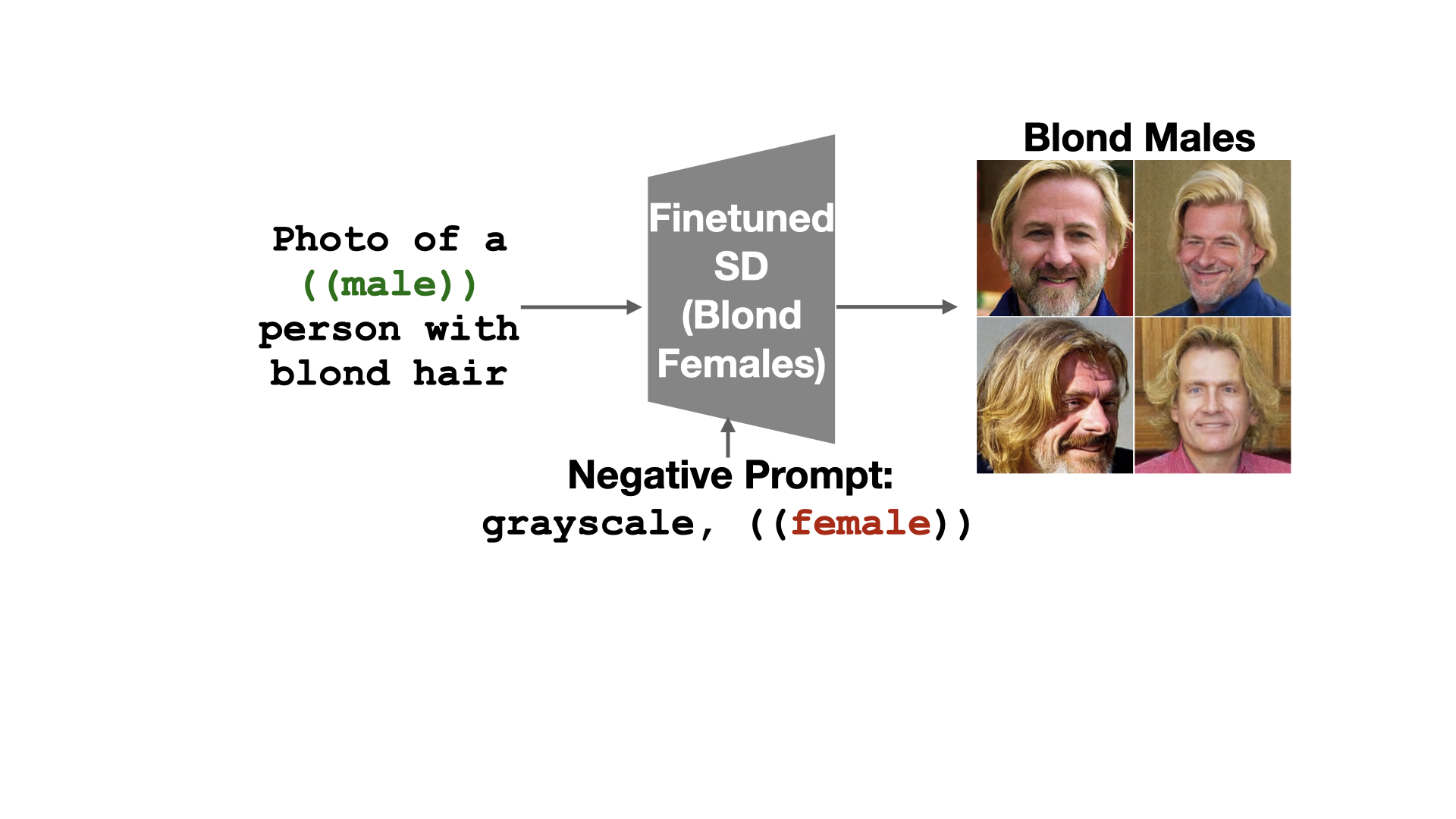}
    \caption{\textbf{Image Generation Pipeline for Bias Ratio$\mathbf{=0.999}$}. Bias-conflicting samples (e.g., Blond Males in CelebA) are generated using models trained on bias-aligned images (e.g., Blond Females).}
    \label{fig:transfer}
\end{figure}

% \noindent \textbf{Textual Inversion (TI).} These models learn a set of tokens to represent a group of images that share a common property or object~\cite{gal2022image}. As each group in the training data has specific properties or attributes (e.g., the ``blond-male'' group in the CelebA dataset consists of male faces with blond hair), we learn distinct tokens for each group in the training data. The text-to-image model is then prompted to generate images for each group using these learned tokens.

% From each of these variants, we generate images from each group uniformly, rank them using the CLIP-Score (eq.~\ref{eq:2}), and choose the top $m$ images from each group.

\noindent \textbf{Varying Bias Severity}. We extend our evaluation to scenarios where each dataset is severely biased, setting the bias ratio to $0.999$ (i.e., $99.9\%$ of training images belong to bias-aligned groups). The diffusion models must generate images for minority groups to counteract bias severity. For Vanilla SD and FFR, images are generated by prompting with the bias label $a$ and class label $y$. As LoRA-finetuned SD, DreamBooth, and Clustered DreamBooth rely on the training group images, the bias-conflicting images are generated using models trained on bias-aligned groups (e.g., Blond Males are generated from the model trained on Blond Females). Interestingly, for the Dreambooth models, we find that during the bias-conflicting sample generation, removing the learnt `[\texttt{V}]' token from the prompt leads to more accurate depiction of the target group descriptions. On manual inspection, we find that this way of transferring the style of one group into another (described in Fig.~\ref{fig:transfer}) leads to images that visually follow the distribution of the input data, while imitating the target group. Generated images are filtered using the CLIP-Label score with $\alpha=1$ (see eq.~\ref{eq:2}), as the minority groups lack sufficient samples for the CLIP-Centroid computation. The classifier is then pretrained on group-balanced synthetic images and finetuned on the severely biased dataset for each method.

\begin{table*}[t!h!]
\centering
\caption{\textbf{Classification Performance} for the Original Dataset and high-bias variant with respect to each finetuning variant on three datasets. While each finetuning method achieves worst group accuracy (WGA) comparable to debiasing methods like GDRO and SELF for the original datasets, they far outperform the latter for the high bias-ratio case. Clustered Dreambooth achieves highest WGA across all datasets ($79.1\%$). Our results are averaged across $3$ random seeds. $\dagger$ indicates implementation using the codebases of existing methods. The best and 2nd best scores are marked in bold and underline respectively.}
\label{tab:main-table}
\small
\resizebox{0.95\columnwidth}{!}{
\begin{tabular}{c|c|c|ll|ll|ll}
\toprule
\multirow{3}{*}{Dataset} & \multirow{3}{*}{Method} & \multirow{3}{*}{\begin{tabular}[c]{@{}c@{}}Synthetic \\ Data?\end{tabular}} &  \multicolumn{2}{c|}{Original Dataset} & \multicolumn{2}{c|}{Bias Ratio 0.999} & \multicolumn{2}{c}{Average Performance} \\
\cmidrule{4-9} 
 &  &  & Worst & Average & Worst & Average & Worst & Average \\
\midrule
\multirow{6}{*}{Waterbirds} & ERM & \XSolid & $63.7$ & $88.0$ & $29.0$ & $66.7$ & $46.3$ & $77.3$ \\
\cmidrule{2-9} 
& FFR$\dagger$~\cite{qraitem2023fake} & \Checkmark &  $69.5$ & $84.0$ & $57.3$ & $84.2$ & $63.4$ & $84.1$ \\
 & Vanilla SD & \Checkmark & $74.6^{\pm 2.90}$ & $80.5^{\pm 0.34}$ & $69.9^{\pm 0.70}$ & $80.1^{\pm 0.13}$ & $72.2$ & $80.3$ \\
 & LoRA-finetuning & \Checkmark & $86.5^{\pm 3.81}$ & $89.9^{\pm 0.76}$ & $61.5^{\pm 0.40}$ & $84.0^{\pm 0.11}$ & $74.0$ & $87.0$ \\
 & Dreambooth~\cite{ruiz2022dreambooth} & \Checkmark &  $\mathbf{89.3}^{\pm 0.75}$ & $\underline{90.1}^{\pm 0.50}$ & $82.4^{\pm 0.25}$ & $\underline{88.3}^{\pm 0.22}$ & $\underline{85.9}$ & $\underline{89.2}$ \\
 & Clustered Dreambooth & \Checkmark &  $88.1^{\pm0.92}$ & $\mathbf{90.2}^{\pm 0.11}$ & $\mathbf{84.2}^{\pm 0.46}$ & $\mathbf{88.5}^{\pm 0.14}$ & $\mathbf{86.0}$ & $\mathbf{89.3}$ \\
\cmidrule{2-9} 
 & GDRO$\dagger$~\cite{sagawa2019distributionally} & \XSolid & $91.4$  & $93.5$ & $23.5$ & $65.5$ & $57.4$ & $79.5$ \\
 & SELF$\dagger$~\cite{labonte2023towards} & \XSolid & $93.0$  & $94.0$ & $25.5$ & $64.2$ & $59.2$ & $79.1$ \\
\midrule
\multirow{6}{*}{CelebA} & ERM & \XSolid &  $47.8$ & $94.9$ & $31.7$ & $67.3$ & $39.7$ & $81.1$ \\
\cmidrule{2-9} 
& FFR$\dagger$~\cite{qraitem2023fake} & \Checkmark &  $68.9$ & $85.7$ & $22.8$ & $47.7$ & $45.9$ & $66.7$ \\
 & Vanilla SD & \Checkmark &  $76.4^{\pm 1.27}$ & $84.2^{\pm 0.37}$ & $77.1^{\pm 0.42}$ & $\underline{84.7}^{\pm 0.67}$ & $76.7$ & $84.4$ \\
 & LoRA-finetuning & \Checkmark &  $\underline{82.3}^{\pm 1.51}$ & $87.2^{\pm 0.56}$ & $73.5^{\pm 2.83}$ & $83.2^{\pm 0.29}$ & $77.9$ & $85.2$ \\
 & Dreambooth~\cite{ruiz2022dreambooth} & \Checkmark &  $82.1^{\pm 0.00}$ & $\underline{87.9}^{\pm 0.32}$ & $\underline{78.8}^{\pm 0.21}$ & $84.6^{\pm 0.19}$ & $\underline{80.4}$ & $\underline{86.2}$ \\
 & Clustered Dreambooth & \Checkmark &  $\mathbf{84.1}^{\pm 0.63}$ & $\mathbf{88.4}^{\pm 0.19}$ & $\mathbf{81.8}^{\pm 0.35}$ & $\mathbf{85.9}^{\pm 0.28}$ & $\mathbf{82.9}$ & $\mathbf{87.1}$ \\
\cmidrule{2-9} 
 & GDRO$\dagger$~\cite{sagawa2019distributionally} & \XSolid  & $88.9$ & $92.9$  & $27.2$ & $75.2$ & $58.0$ & $84.0$ \\
 & SELF$\dagger$~\cite{labonte2023towards} & \XSolid  & $83.9$ & $91.1$  & $45.6$ & $95.4$ & $64.7$ & $93.2$ \\
\midrule
\multirow{6}{*}{UTKFace} & ERM & \XSolid &  $74.3$ & $84.5$ & $31.0$ & $48.9$ & $52.6$ & $66.7$ \\
\cmidrule{2-9} 
& FFR$\dagger$~\cite{qraitem2023fake} & \Checkmark &  $67.4$ & $81.4$ & $55.0$ & $68.0$ & $61.2$ & $74.7$ \\
 & Vanilla SD & \Checkmark &  $62.0^{\pm 3.89}$ & $83.3^{\pm 0.92}$ & $\mathbf{67.8}^{\pm 1.27}$ & $\mathbf{82.7}^{\pm 0.61}$ & $64.9$ & $\underline{83.0}$ \\
 & LoRA-finetuning & \Checkmark &  $68.6^{\pm 3.91}$ & $\mathbf{85.6}^{\pm 0.76}$ & $64.5^{\pm 2.59}$ & $\underline{82.4}^{\pm 0.26}$ & $\underline{66.5}$ & $\mathbf{84.0}$\\
 & Dreambooth~\cite{ruiz2022dreambooth} & \Checkmark &  $57.9^{\pm 3.91}$ & $80.9^{\pm 0.76}$ & $\underline{66.9}^{\pm 2.59}$ & $77.1^{\pm 0.26}$ & $62.4$ & $79.0$\\
 & Clustered Dreambooth & \Checkmark &  $\mathbf{76.0}^{\pm 1.22}$ & $\underline{83.5}^{\pm 0.35}$ & $60.5^{\pm 1.22}$ & $80.8^{\pm 0.35}$ & $\mathbf{68.2}$ & $82.1$ \\
 \cmidrule{2-9} 
 & GDRO$\dagger$~\cite{sagawa2019distributionally} & \XSolid & $81.6$ & $85.9$ & $30.5$ & $50.3$ & $56.0$ & $68.1$ \\
 & SELF$\dagger$~\cite{labonte2023towards} & \XSolid & $65.9$ & $82.3$ & $0.6$ & $50.5$ & $33.3$ & $66.4$ \\
\bottomrule
\end{tabular}}
\vspace{-0.2cm}
\end{table*}

% \vspace{-0.2cm}
\noindent \textbf{Implementation Details}:
% \vspace{-0.1cm}
We use SD v1.4 for all experiments. The prompts used to generate the images for each variant is provided in Appendix~\ref{subsec:prompts}. For each dataset group, we generate $M = 5000$ images for each method, and score the images using the CLIP-Score defined in eq.~\ref{eq:2}. For simplicity, we set the weighting parameter $\alpha$ to be $0.5$ for the original dataset ($\alpha=1$ for the severely biased version). Accordingly, we select the top $75\%$ images from each group. An ImageNet~\cite{deng2009imagenet}-pretrained ResNet-50~\cite{He_2016_CVPR} model is trained on the group-balanced data with CE and SupCon losses. The model is trained for 20 epochs using the SGD optimizer during both the training and finetuning stages. As the validation sets of the datasets are not group-balanced, we refrain from experimenting with hyperparameters, and uniformly set a learning rate of $1\text{e}-3$, weight decay of $1\text{e}-3$ and batch size of $128$. We discuss the choice of clusters for the Clustered Dreambooth method in \S~\ref{subsec: design_choices_main}.
% We refer the reader to the supplementary material for more details on our design choices.

% \vspace{-0.1cm}
\subsection{Quantitative Results}
% \vspace{-0.1cm}
Our goal is to obtain rich representations from synthetic images to enable classification models to be fair, even when finetuned on biased real data.
%Our goal is not to achieve state-of-the-art results, rather we aim to show that rich representations from synthetic images enable classification models to be relatively fair, even when finetuned on biased real data. 
% We also report the accuracies of FFR~\cite{qraitem2023fake}, which generates images using vanilla Stable Diffusion. 
Table~\ref{tab:main-table} presents the worst and average group accuracies (WGA and AGA) across all test groups and dataset variants. To further motivate the importance of the generative models in training fair classifiers, we compare their performance with traditional debiasing methods like Group-DRO (GDRO)~\cite{sagawa2019distributionally} and SELF~\cite{labonte2023towards}. Key findings are summarized below:

\noindent \textbf{Performance on the Original Dataset}. Our classification pipeline, leveraging images from vanilla SD, outperforms FFR~\cite{qraitem2023fake} by an average of $2.4\%$. This improvement can be attributed to our last layer retraining-based finetuning stage. DreamBooth achieves the highest worst-group accuracy for Waterbirds ($89.3\%$), followed by Clustered DreamBooth ($88.1\%$) and LoRA fine-tuning ($86.5\%$). For CelebA and UTKFace, Clustered DreamBooth outperforms all other generative methods and is the only one to surpass ERM scores for UTKFace, improving upon LoRA by $7.4\%$. This highlights the advantage of training multiple DreamBooth models on clustered subsets, particularly for facial datasets. While Clustered DreamBooth has higher time complexity, $k_D$ can be adjusted based on resource constraints, with $k_D=1$ (a single DreamBooth model per group) as a special case. Notably, since DreamBooth models are themselves trained with LoRA, per-model training remains computationally efficient. Overall, fine-tuning consistently outperforms vanilla Stable Diffusion across all datasets, achieving accuracy comparable to traditional debiasing methods like Group-DRO.

\noindent \textbf{Performance for Bias Ratio $\mathbf{=0.999}$}. With severe bias ratio, traditional debiasing methods experience a drastic performance drop, whereas generative methods exhibit significantly lower degradation.
For Waterbirds and CelebA, DreamBooth and Clustered DreamBooth outperform all other methods, with performance gaps between the original and biased versions remaining below $7\%$, compared to over $60\%$ for traditional debiasing approaches. This emphasizes the importance of leveraging generative models for training fairer classifiers. Additionally, for these datasets, the accuracies of vanilla SD and LoRA finetuning remain similar. 
For UTKFace, all generative finetuning methods underperform compared to the vanilla SD pipeline, which achieves the highest accuracy. Manual inspection reveals many bias-conflicting samples are irrelevant or out-of-domain (see Appendix Fig.~\ref{fig:utkfailure}). We leave further investigation and mitigation of this problem for future work. Averaged across all datasets, fine-tuning approaches outperform vanilla Stable Diffusion, with Clustered Dreambooth achieving $75.5\%$ WGA.

\noindent \textbf{Generative-Based Pretraining + GDRO Finetuning ($\mathbf{0.999}$ bias ratio)}. Table~\ref{tab:main-table} shows that the WGA of GDRO~\cite{sagawa2019distributionally} suffers for high bias ratio. We next analyse the impact of finetuning the classification model pretrained by the Clustered Dreambooth images~\footnote{We choose Clustered Dreambooth for different ablations and experiments only as an example.} for Waterbirds and CelebA, using the GDRO loss, leading to a $58.3\%$ and $37.2\%$ performance increase for Waterbirds and CelebA respectively (Table~\ref{tab:gdro}). This analysis further highlights the positive effect of group-balanced synthetic image pretraining. Qualitative examples are shown in Appendix Fig.~\ref{fig:999}.

% Please add the following required packages to your document preamble:
% \usepackage{booktabs}
% \usepackage{multirow}

% Please add the following required packages to your document preamble:
% \usepackage{booktabs}
% \usepackage{multirow}

\begin{table}
    \centering
    \small
    \caption{\textbf{Clustered Dreambooth + GDRO} for Waterbirds and CelebA: i.e., pretrain the classification model using the synthetic images, and then finetune the same using GDRO.}
    \label{tab:gdro}
    \resizebox{0.8\linewidth}{!}{%
        \begin{tabular}{@{}l|ll|lc@{}}
        \toprule
        \multirow{2}{*}{Method} & \multicolumn{2}{c|}{GDRO} & \multicolumn{2}{c}{\begin{tabular}[c]{@{}c@{}}Clustered Dreambooth \\ (Pretraining)+ GDRO\end{tabular}} \\ \cmidrule(l){2-5} 
         & \multicolumn{1}{l|}{WGA} & AGA & \multicolumn{1}{l|}{WGA} & AGA \\ \midrule
        Waterbirds & \multicolumn{1}{l|}{$23.5$} & $65.5$ & \multicolumn{1}{l|}{$81.8$} & $89.8$ \\
        CelebA     & \multicolumn{1}{l|}{$27.2$} & $75.2$ & \multicolumn{1}{l|}{$64.4$} & $85.0$ \\ 
        \bottomrule
        \end{tabular}%
    }
    % \vspace{-0.35cm}
\end{table}

\noindent \textbf{Time Complexity of Finetuning}. We analyze the time complexity vs performance tradeoff for the finetuning techniques, averaged across all datasets and bias ratios. Clustered DreamBooth outperforms others, but it has higher time complexity of $\mathcal{O}(|\mathcal{G}_D|.k_D)$, where $|\mathcal{G}_D|$ represents the number of groups in the training set, and $k_D:$ the number of clusters per group. In contrast, vanilla SD, which requires no finetuning, yields the lowest score across all datasets. These tradeoffs are presented in Table~\ref{tab:time-complexity}, providing insights to help practitioners select the most suitable method based on their application needs. We also evaluate distribution similarities using Fréchet Inception Distance (FID)~\cite{heusel2017gans} between generated and real CelebA images: finetuning approaches outperform FFR significantly (see Appendix Table~\ref{tab:fid}).

\subsection{Design Choices}
% \vspace{-0.2cm}
\label{subsec: design_choices_main}

\begin{figure*}[h!]
    \centering
    \includegraphics[width=0.8\textwidth, trim=0 100 0 100, clip]{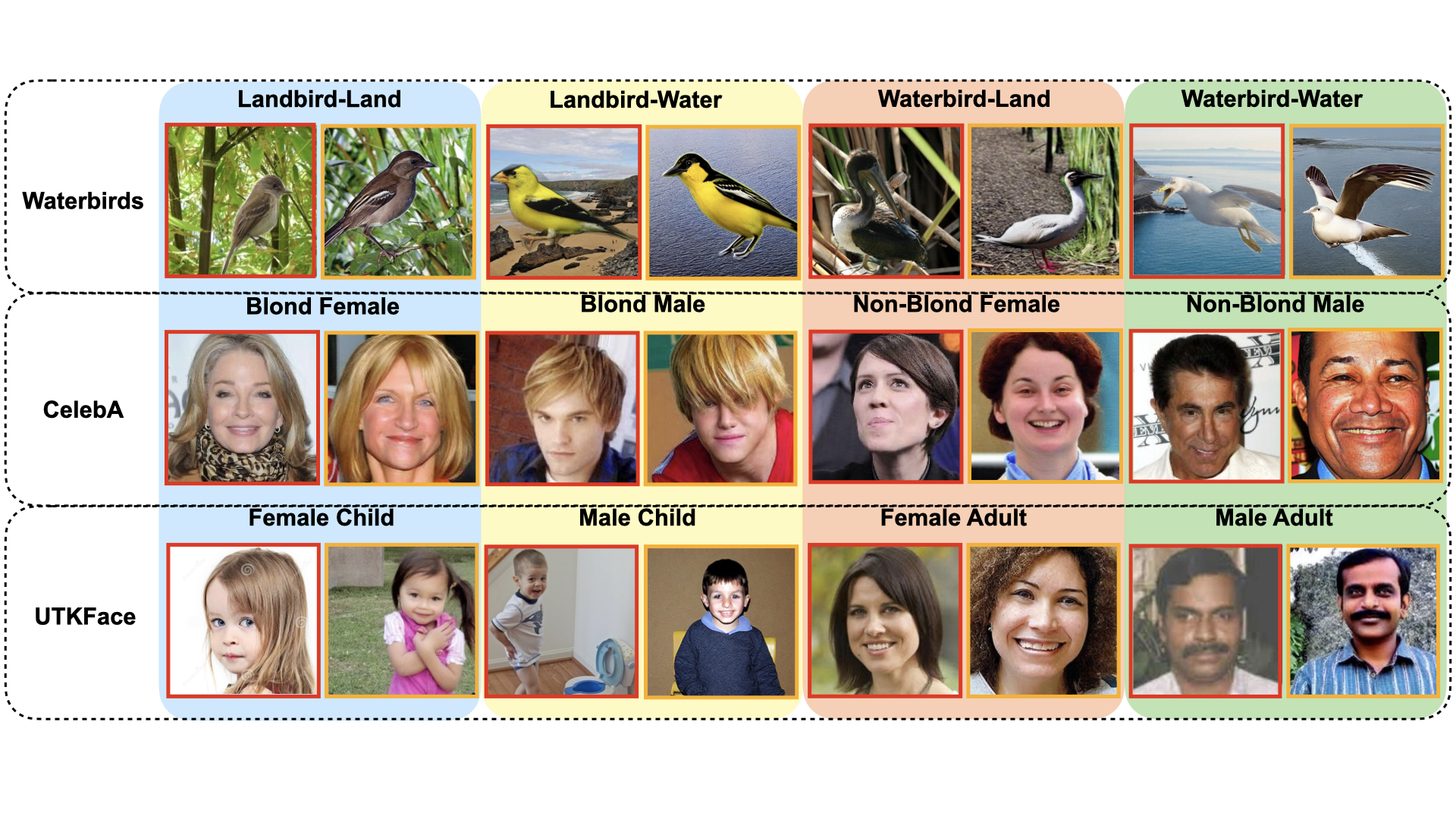}
    \caption{\textbf{Training vs Generated Images from Clustered Dreambooth}. In this figure, we present original images from each group in the studied datasets (images with red border) along with the generated ones (yellow border). We note how the generated images closely reflect the training distribution and the group-related attributes.}
    \label{fig:qualitative}
    \vspace{-0.3cm}
\end{figure*}

\begin{figure}[h!]
    \centering
    \includegraphics[width=0.9\linewidth, trim=20 420 60 355, clip]{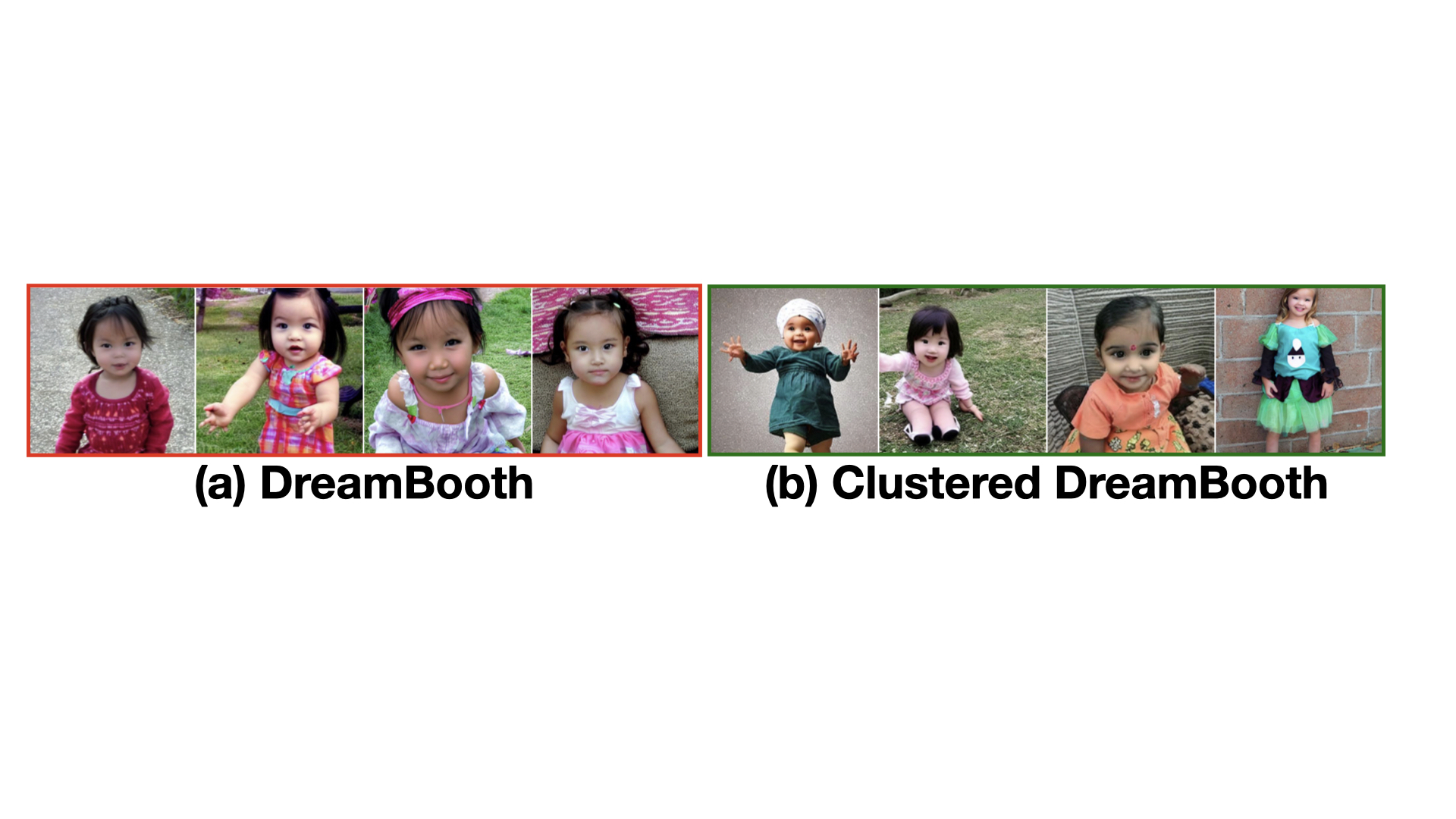}
    \caption{\textbf{Dreambooth vs Clustered Dreambooth for UTKFace Female Children.} Clustered Dreambooth samples are more diverse than the Dreambooth ones.}
    \label{fig:utkface}
\end{figure}

\textbf{Choice of Clusters for Clustered Dreambooth}.
We select the number of clusters $k_D$ for Clustered DreamBooth based on the size of the smallest group in the dataset, $M_{g_s}$. To ensure sufficient training data per cluster, we set $k_D \approx \frac{M_{g_s}}{20}$, keeping at least 20 samples per cluster. To limit complexity on larger datasets, we cap the number of clusters: $k_D = \min\left(\frac{M_{g_s}}{20}, 20\right)$. Thus, we use $k_D = 3$ for Waterbirds ($M_{g_s} = 56$), $k_D = 20$ for CelebA ($M_{g_s} = 1387$), and $k_D = 5$ for UTKFace ($M_{g_s} = 103$). For simplicity, $k_D$ is fixed per dataset and shared across all groups, assuming similar intra-group variation. Further optimization of $k_D$ is left for future work.

\noindent \textbf{Group-Balanced Finetuning}.
After pretraining on generated data, we finetune the classification layer on real data (LLR$_{\text{all}}$). To assess the benefits of group-balancing the real data (sized to the smallest group), we explore:
a) \textit{Last Layer Retraining with Balanced Real Data} (LLR$_{\text{b}}$): Finetuning only the classification layer using the balanced real dataset instead of the full biased set.
b) \textit{Full Fine-Tuning with Balanced Real Data} (FT$_{\text{b}}$): Finetuning the entire network with the balanced real dataset.
Results show that FT$_{\text{b}}$ is beneficial only for CelebA, likely due to its larger size compared to Waterbirds and UTKFace (Appendix Table~\ref{tab:group-balance}). In contrast, LLR$_{\text{b}}$ reduces worst group accuracy across all datasets, indicating that finetuning on the entire dataset yields better performance.

\begin{table}
    \centering
    \small
    \caption{\textbf{Classification Performance vs Time Complexity Tradeoff}, averaged across all datasets and both bias ratios.}
    \label{tab:time-complexity}
    \resizebox{0.8\linewidth}{!}{%
        \begin{tabular}{@{}l|c|l|l@{}}
        \toprule
        Method & Time Complexity & WGA & AGA \\ \midrule
        Vanilla SD & $\mathcal{O}(1)$ & $71.3$ & $82.6$ \\
        LoRA-finetuning & $\mathcal{O}(|\mathcal{G}_D|)$ & $72.8$ & $85.4$ \\
        Dreambooth & $\mathcal{O}(|\mathcal{G}_D|)$ & $76.2$ & $84.8$ \\
        Clustered Dreambooth & $\mathcal{O}(|\mathcal{G}_D|.k_D)$ & $\mathbf{79.1}$ & $\mathbf{86.2}$ \\ 
        \bottomrule
        \end{tabular}%
    }
    \vspace{-0.3cm}
\end{table}

We conduct additional ablations on our design choices: (i) \textit{Role of CE and SupCon losses}: The SupCon loss plays a crucial role in improving fairness when training on severely biased dataset versions (Appendix Tables~\ref{tab:wb-supcon} and \ref{tab:ut-supcon}). (ii) \textit{Single-stage vs. two-stage training}: Training a classifier in a single stage by combining synthetic and real data leads to performance drop, likely due to domain mismatches (Appendix Table~\ref{tab:usb}), (iii) We provide a detailed hyperparameter analysis in Appendix Fig.~\ref{fig:hyp}.

% \vspace{-0.2cm}
\subsection{Qualitative Analysis}
% \vspace{-0.2cm}
\label{subsec: qualitative}
\noindent \textbf{Clustered Dreambooth vs Original}. Figure~\ref{fig:qualitative} shows original images alongside synthetic ones for each data group, showing strong alignment with group characteristics and domain preservation, leading to high WGA even during pretraining. Fig.~\ref{fig:celeba} (Appendix) shows different characteristics represented by images from different clusters in CelebA.

\noindent \textbf{Dreambooth for UTKFace}. Table~\ref{tab:main-table} shows that the performance of models trained on vanilla Dreambooth drops in case of UTKFace (original). To investigate, we present \textit{female children} faces in Figure~\ref{fig:utkface}. The Dreambooth images follow a narrow demographic and age range, whereas the Clustered Dreambooth images are more diverse in demographics, age groups, and backgrounds. This shows Clustered Dreambooth’s strength in capturing variety within training groups, unlike a single model trained on all group images. More examples on the images generated from different clusters are shared in Appendix~\ref{fig:cluster}.

% \begin{wraptable}{r}{0.58\columnwidth}
%     \centering
%     \small
%     \caption{\textbf{Effect of Scale} (i.e., number of images generated). We evaluate Waterbirds performance using Dreambooth and Clustered Dreambooth for 20K to 5K generated images (top $75\%$ retained). No significant performance drop is observed.}
%     \label{tab:scale}
%     \resizebox{\linewidth}{!}{%
%         \begin{tabular}{@{}l|ll|cc@{}}
%         \toprule
%         \multirow{2}{*}{Scale} & \multicolumn{2}{l|}{Dreambooth} & \multicolumn{2}{l}{Clustered Dreambooth} \\ \cmidrule(l){2-5} 
%          & \multicolumn{1}{l|}{WGA} & AGA & \multicolumn{1}{l|}{WGA} & AGA \\ \midrule
%         20K & \multicolumn{1}{l|}{$88.3$} & $90.3$ & \multicolumn{1}{l|}{$88.8$} & $90.4$ \\
%         15K & \multicolumn{1}{l|}{$88.5$} & $90.5$ & \multicolumn{1}{l|}{$87.9$} & $90.4$ \\
%         10K & \multicolumn{1}{l|}{$88.2$} & $89.9$ & \multicolumn{1}{l|}{$87.9$} & $90.25$ \\
%          5K & \multicolumn{1}{l|}{$89.3$} & $90.1$ & \multicolumn{1}{l|}{$87.8$} & $90.2$ \\ 
%         \bottomrule
%         \end{tabular}%
%     }
% \end{wraptable}

% Please add the following required packages to your document preamble:
% \usepackage{booktabs}
% \usepackage{multirow}

\vspace{-0.1cm}
\section{Conclusion}
\label{sec: conclusion}
\vspace{-0.1cm}
In this work, we investigated Stable Diffusion and various finetuning approaches, including Dreambooth and LoRA, to enhance fairness in image classification by generating representative images for each training group. We also explored Clustered Dreambooth, which addresses intra-group diversity by clustering images within each group and training separate Dreambooth models per cluster, preventing a single model from being overwhelmed by excessive variation. Using these approaches, we generated group-balanced images, pretrained a classifier on them, and finetuned it on real data. Experiments on three fairness benchmarks demonstrated that diffusion-based finetuning, particularly Clustered Dreambooth, consistently outperforms vanilla SD and FFR~\cite{qraitem2023fake}, achieving comparable or superior results to SOTA debiasing methods like GroupDRO—especially as dataset biases become more severe.

\noindent \textbf{Limitations}. Our work has some limitations. Unlike vanilla SD, finetuned variants require dedicated training, with image quality sensitive to hyperparameters. For Clustered Dreambooth, optimizing the cluster count $k_D$ through further experiments could improve performance. Additionally, generated images alone cannot fully mitigate biases, necessitating a two-stage approach (see Appendix Table~\ref{tab:usb}). Despite these challenges, our study highlights the potential of diffusion models for enhancing fairness in classification tasks.

\appendix
\appendix
\section{Appendix}
We organize this appendix as follows. We begin with details on the loss functions used to train our classification pipeline and the data split for the UTKFace dataset~\cite{zhang2017age}. Next, we describe the prompts used for generating images from different diffusion model variants. We then present key ablations on our design choices, where we demonstrate the importance of the CLIP-Score introduced in eq.~\ref{eq:2}, analyze various strategies for the classification finetuning stage, and examine the impact of selecting different proportions of synthetic images for pretraining. Additionally, we discuss the roles of CE and SupCon losses in the classification pipeline. We conclude with an analysis of hyperparameter effects during pretraining with group-balanced synthetic data, an evaluation of representation similarity between Clustered Dreambooth images and real data, a report on worst-group classification performance after the pretraining stage, and qualitative examples of images generated by different methods explored in this work.

% \subsection{Experiments - Extended Details}

\subsection{Loss Function for the Two-Stage Pipeline}
\label{subsec: losses}
Recall that our two stage pipeline is trained by a weighted combination of the Cross-Entropy (CE) loss and the Supervised Contrastive (SupCon) loss. We define both of them formally here. Given the set of training images $\mathcal{X}$, where each $x_i \in \mathcal{X}$ is associated with a class label $y_i \in \mathcal{Y}$, the CE loss is defined as:

\begin{equation}
\label{eq:1}
    L_{\text{CE}} = \frac{1}{|\mathcal{X}|}\sum_{i=1}^{|\mathcal{X}|}\sum_{j=1}^{|\mathcal{Y}|}-p_{ij}\log\hat{p}_{ij}
\end{equation}
where $[p_{i1}, p_{i2}, \cdots, p_{i|\mathcal{Y}|}]$ is the one-hot vector representation of $y_i$ and $[\hat{p}_{i1}, \hat{p}_{i2}, \cdots, \hat{p}_{i|\mathcal{Y}|}]$ is the corresponding softmax vector, obtained from the model classification model $f$ (see \S~\ref{subsec: gen_role} in the main paper).

The SupCon loss encourages the model to learn more discriminative features by promoting greater separation between samples from different classes. We consider a mini-batch of size $\mathcal{B}$ of features denoted as $\{\mathbf{e(x_1)}, \mathbf{e(x_2)}, \cdots, \mathbf{e(x_\mathcal{B})}\}$ and corresponding class labels as $\{y_1, y_2, \dots, y_\mathcal{B}\}$ ($e$ is the feature encoder). 
% %Since multiple answers are possible for a single image-question pair, we choose the most probable one for computing the loss value. 
Let us consider the current sample with index $j$, and the set of positive examples from the mini-batch by $P_j: \{i \in \mathcal{B} \quad s.t. \quad y_i = y_j\}$. Similarly, the set of negative examples is denoted by $N_j: \{i \in \mathcal{B} \quad s.t. \quad y_i \neq y_j\}$. The SupCon loss \cite{khosla2020supervised} for a image $x_j$ is defined as,
\begin{equation}
    L_{\text{sup-con}} = \sum_{j \in \mathcal{B}}\frac{-1}{|P_j|}\sum_{p \in P_j} \log \frac{\exp(\mathbf{e(x_j)}^T \mathbf{e(x_p)}/\tau)}{\sum \limits_{n \in N_j}\exp(\mathbf{e(x_j)}^T \mathbf{e(x_n)}/\tau)},
\end{equation}
We set the temperature $\tau = 1$ for all experiments as we do not find any significant improvements by changing its value.
The final loss function becomes a combination of the CE loss and SupCon loss:
$$L = \beta. L_{\text{CE}} + (1-\beta) L_{\text{sup-con}}$$
where $\beta = 0.5$ for all experiments.

\subsection{UTKFace Splits}
\label{subsec:utk}
Following previous works~\cite{qraitem2023bias, qraitem2023fake}, we deliberately introduce biases into the UTKFace dataset~\cite{zhang2017age}, wherein we use Gender as the target attribute and Age as the bias attribute with $90\%$ bias ratio. Particularly, we ensure that the number of images in the training set for the different groups are: $103$, $934$, $5730$ and $636$ for male adults, male children, female adults and female children respectively. To binarize the Age attribute, ages $\leq 10$ are considered as children, and those $\geq 20$ are considered adults~\cite{qraitem2023bias, qraitem2023fake}. Thus, the females are biased towards older age, whereas the males are biased towards the children.

\subsection{Prompts used for Generation}
\label{subsec:prompts}
Here we present the prompts we used for each method and dataset to generate images from every group. 

\begin{table*}[h!t!]
\centering
\caption{\textbf{Effect of CLIP-Score weighting parameter $\mathbf{\alpha}$} on Clustered Dreambooth final worst group accuracies. We observe that setting $\alpha=1$ outperforms $\alpha=0$, highlighting the effectiveness of CLIP based similarity between the textual form of the class label and the images. Setting $\alpha=0.5$ (as reported in the main paper) works best for most datasets. Randomly selecting the images without using any scoring functions is also seen to perform on par with the other settings, the performance is generally weaker. All scores are with respect to the original versions of the datasets.}
\label{tab:alpha}
\begin{tabular}{@{}l|ll|ll|ll@{}}
\toprule
\multirow{2}{*}{\begin{tabular}[c]{@{}l@{}}Selection\\ Method\end{tabular}} & \multicolumn{2}{l|}{Waterbirds} & \multicolumn{2}{l|}{CelebA} & \multicolumn{2}{l}{UTKFace} \\ \cmidrule(l){2-7} 
 & \multicolumn{1}{l|}{WGA} & AGA & \multicolumn{1}{l|}{WGA} & AGA & \multicolumn{1}{l|}{WGA} & AGA \\ \midrule
$\alpha=1$ & \multicolumn{1}{l|}{$87.0$} & $89.9$ & \multicolumn{1}{l|}{$83.8$} & $87.9$ & \multicolumn{1}{l|}{$72.7$} & $84.5$ \\
$\alpha=0$ & \multicolumn{1}{l|}{$84.7$} & $90.15$ & \multicolumn{1}{l|}{$83.3$} & $87.95$ & \multicolumn{1}{l|}{$68.8$} & $82.9$ \\
$\alpha=0.5$ & \multicolumn{1}{l|}{$\mathbf{88.1}$} & $90.2$ & \multicolumn{1}{l|}{$\mathbf{84.1}$} & $\mathbf{88.4}$ & \multicolumn{1}{l|}{$\mathbf{76.0}$} & $83.5$ \\
Random sampling & \multicolumn{1}{l|}{$84.9$} & $\mathbf{90.4}$ & \multicolumn{1}{l|}{$81.1$} & $87.0$ & \multicolumn{1}{l|}{$73.0$} & $\mathbf{84.62}$ \\ \bottomrule
\end{tabular}
\end{table*}

\noindent \textbf{Prompts for Vanilla SD}~\cite{ramesh2022hierarchical}. For Waterbirds, the prompt used is ``\texttt{photo of a \{class-label\} on \{bias-label\}.}'', where class-label can be landbird or waterbird, and bias-label can be land or water. For UTKFace, the prompt used is ``\texttt{photo of a \{class-label\} \{bias-label\}.}'', where class-label can be female or male, bias-label can be child or adult. Finally, for CelebA, we use the following template: ``\texttt{photo of a \{bias-label\} person with blond hair}'' to generate males and females (i.e., the bias-labels) with blond hair. For the non-blond class, we prompt the model with ``\texttt{photo of a \{bias-label\} person}'', whereas we use a negative prompt having `\texttt{blond hair}', to force the model to generate non-blond males and females. As the vanilla SD is not finetuned on our training sets, we use the same set of prompts for the bias ratio $0.999$ case.

\noindent \textbf{Prompts for Finetuned Diffusion Models}. For LoRA-finetuning, we use prompts of the format `\texttt{Photo of a \{class-label\} on \{bias-label\}}' for Waterbirds (\texttt{class-label} $\in$ \texttt{\{waterbird, landbird\}}, \texttt{bias-label} $\in$ \texttt{\{water, land\}}), and `\texttt{Photo of a \{class-label\} person who is a \{bias-label\}}' for UTKFace (\texttt{class-label} $\in$ \texttt{\{male, female\}}, \texttt{bias-label} $\in$ \texttt{\{adult, child\}}). For CelebA, the prompts used are: `\texttt{Photo of a non-blond \{bias-label\} person}' for the class Non-Blond, and `\texttt{Photo of a \{bias-label\} person with blond hair}' for the class Blond, where \texttt{bias-label} $\in$ \texttt{\{male, female\}}. We use the same set of prompts for Dreambooth~\cite{ruiz2022dreambooth} (i.e., single model per group) and Clustered Dreambooth. Recall that each of these methods learn specific tokens to represent the groups (or clusters in the groups in case of Clustered Dreambooth). We denote the learnt tokens by `\texttt{[V]}'. Likewise, for Waterbirds, we use the prompt ``\texttt{photo of a [V] bird}'', where `\texttt{[V]}' represents the learnt tokens by the model trained on the specific group or cluster. For UTKFace and CelebA, we find that providing the class-label in the prompt generates more accurate images. Hence, for UTKFace, the prompt is of the form ``\texttt{photo of a [V] \{class-label\} person}''. For CelebA, the blond-class images are generated using ``\texttt{photo of a [V] person with blond hair}'', whereas for the non-blond class, the prompt is ``\texttt{photo of a [V] person}'' with blond hair as the negative prompt.

\noindent \textbf{Prompts for Bias Ratio $\mathbf{=0.999}$ Scenario}. The prompts used are similar to the case of the original dataset for each generation method. For UTKFace, we generate the female children faces from model(s) trained on the male children faces, and male adult faces from those trained on the female adult faces. To enforce the model to generate correct images from the bias-conflicting groups, we emphasize the \texttt{class-label} in the prompt by placing it inside double parenthesis. We also add the opposite class-label to the negative prompt with double parenthesis. For example, the generation prompt for female children is `\texttt{Photo of a ((female)) person who is an child}', with an additional negative prompt `\texttt{((male))}''. For CelebA, we generate blond males from blond female models, and non-blond females from non-blond male models. For Waterbirds, we generate landbird on water images from landbird on land models, whereas waterbird on land images are generated from waterbird on water images. Similar to the UTKFace case, we put double parenthesis in the prompts, but on the \texttt{bias-label} for these two datasets, with the opposite bias-label added to the negative prompts.

\subsection{Extended Ablation Studies}
\label{subsec: ablation2}
Here we present further ablation studies of the proposed approach as an extension to \S~\ref{subsec: design_choices_main} in the main paper. 

\noindent \textbf{Role of The CLIP-Score}. We have described the CLIP-Score in eq.~\ref{eq:2} (main paper, subsection~\ref{subsec: finetune}), used to filter the best $75\%$ images out of the generated ones for each group. We vary the weighting parameter $\alpha$ to understand the role of the label-based score function CLIP-Label($I, p^c$) and the group-centroid based score function CLIP-Centroid($I, \bar{z}^g$). Setting $\alpha=1$ denotes that the scoring function is only dependent on CLIP-Label, whereas $\alpha=0$ denotes otherwise. We present the results of these variants on the Clustered Dreambooth pipeline in Table~\ref{tab:alpha} (with respect to the original dataset versions). Recall that the numbers reported in the main paper correspond to $\alpha=0.5$. We also show a baseline where from each group, images are selected randomly instead of ranking them using the scoring function. We observe that the performance of the pretrained model trained on the images chosen by setting $\alpha=1$ always outperforms its counterpart trained on images selected by setting $\alpha=0$. However, in general, our choice of $\alpha=0.5$ works best across datasets. Random selection also performs on par with the other variants, which shows that the generated images are mostly useful in training fairer classifiers, however, their performances are lower than those involving image selection with the CLIP-Score.

% Please add the following required packages to your document preamble:
% \usepackage{booktabs}
% \usepackage{multirow}

\noindent \textbf{Direct Combination of Real and Synthetic Data}. We observe two settings: a) \textit{Real+Group-Balanced Synthetic} Images: Combine the entire real data and the group-balanced synthetic images generated using Clustered Dreambooth, b) \textit{Group-Balanced (Real+Synthetic) Images}: Combine the real and the synthetic data in such a way that the final images are group-balanced. For both settings, there is only a single stage of training, with the combination of real and synthetic images. Using the images obtained from each stage, we train a classification model, and compare their performances with the pretraining and finetuning stage of the Clustered Dreambooth pipeline. Our experiments show that for Waterbirds and CelebA, the performance drops for both the settings compared to the Clustered Dreambooth pretraining and finetuning stages. UTKFace is an exception, where the worst group-accuracy is high for the \textit{Real+Group-Balanced Synthetic} setting. However, the average group accuracy drops, showing that the accuracy of the groups other than the worst group remains low. Overall, we observe that our two stage approach is considerably more effective than the single stage alternatives (Table~\ref{tab:usb}), justifying our choice in the main paper.

\begin{table*}[]
\centering
\caption{\textbf{Classification Performance on Different Combinations of Real and Synthetic Data}. We evaluate classification systems for two cases: a) The entire real data + group-balanced synthetic images, b) Combination of real and synthetic data ensuring that the resultant dataset is group-balanced. These cases are compared against the pretraining and finetuning stages reported in the main paper for the Clustered Dreambooth pipeline. The experiment shows the two observed cases to be inadequate compared to our pipeline of pretraining with synthetic data and finetuning with real data.}
\label{tab:usb}
\begin{tabular}{@{}l|ll|ll|ll@{}}
\toprule
\multirow{2}{*}{Selection Method} & \multicolumn{2}{l|}{Waterbirds} & \multicolumn{2}{l|}{CelebA} & \multicolumn{2}{l}{UTKFace} \\ \cmidrule(l){2-7} 
 & \multicolumn{1}{l|}{WGA} & AGA & \multicolumn{1}{l|}{WGA} & AGA & \multicolumn{1}{l|}{WGA} & AGA \\ \midrule
Real+Group-Balanced Synethic Images & \multicolumn{1}{l|}{$77.9$} & $88.4$ & \multicolumn{1}{l|}{$48.3$} & $83.2$ & \multicolumn{1}{l|}{$74.8$} & $82.5$ \\
Group-Balanced (Real+Synthetic) Images & \multicolumn{1}{l|}{$79.1$} & $88.5$ & \multicolumn{1}{l|}{$46.7$} & $82.4$ & \multicolumn{1}{l|}{$70.4$} & $82.0$ \\
% Group-Balanced Synthetic Images (Pretraining Stage) & \multicolumn{1}{l|}{$86.2$} & $88.7$ & \multicolumn{1}{l|}{$83.0$} & $88.0$ & \multicolumn{1}{l|}{$67.8$} & $84.5$ \\
Real Images (Finetuning Stage) & \multicolumn{1}{l|}{$88.1$} & $90.2$ & \multicolumn{1}{l|}{$84.1$} & $88.4$ & \multicolumn{1}{l|}{$76.0$} & $83.5$ \\
\bottomrule
\end{tabular}
\end{table*}

\noindent \textbf{Effect of Selection Percentage of Generated Images}.
In this subsection, we study the effect of selecting different percentages of top ranked images as per the CLIP-Score defined in eq.~\ref{eq:2} (Section~\ref{subsec: finetune}, main paper) on Waterbirds (both original and the severely biased variants) for Clustered Dreambooth. Investigating the classifier performance based on top $100\%$, $75\%$ and $50\%$ generated images, across both the dataset variants, we find that selecting the top $75\%$ of the synthetic images appears to be more beneficial for performance, though we do not observe a drastic fall in scores with the other percentages as well. The results are shown in Table~\ref{tab:percentage}.

\begin{table}[]
\centering
\small
\caption{\textbf{Percentage of Synthetic Images selected for Stage 1 Training vs Performance} for Clustered Dreambooth with respect to Waterbirds. We find that across the moderate and severe bias ratios, while scores do not vary drastically with selection percentages, it is beneficial to choose the top $75\%$ images, as it leads to better performance than the other percentages across both the dataset variants.}
\label{tab:percentage}
\begin{tabular}{l|ll|ll}
\hline
\multirow{2}{*}{\begin{tabular}[c]{@{}l@{}}Selection\\ Percentage\end{tabular}} & \multicolumn{2}{c|}{Clustered Dreambooth} & \multicolumn{2}{c}{Bias Ratio 0.999} \\ \cline{2-5} 
 & \multicolumn{1}{l|}{WGA} & AGA & \multicolumn{1}{l|}{WGA} & AGA \\ \hline
$50\%$ & \multicolumn{1}{l|}{$86.4$} & $89.8$ & \multicolumn{1}{l|}{$82.5$} & $88.1$ \\ 
$75\%$ & \multicolumn{1}{l|}{$\mathbf{88.1}$} & $\mathbf{90.2}$ & \multicolumn{1}{l|}{$\mathbf{84.2}$} & $\mathbf{88.5}$ \\ 
$100\%$ & \multicolumn{1}{l|}{$\mathbf{88.1}$} & $90.0$ & \multicolumn{1}{l|}{$81.8$} & $87.2$ \\ \hline
\end{tabular}
\end{table}

\noindent \textbf{Effect of the Loss Functions in Stage 1 and Stage 2 Training}. To train the classification model in the synthetic image pretraining stage and real image finetuning stage, we use a weighted combination of the CE and SupCon losses. Here, we show the importance of combining these losses in both stages, by demonstrating their effects on the performance of Clustered Dreambooth for Waterbirds and UTKFace, in case of both the original dataset and the severely biased variant (see Tables~\ref{tab:wb-supcon} and \ref{tab:ut-supcon} respectively). We find that while the effect of the SupCon loss is less pronounced for the original versions of both the datasets, its impact is clearly visible for the severely biased versions, especially for UTKFace. Recall that for Clustered Dreambooth, the images of the bias-conflicting samples are generated from the diffusion models finetuned on the bias-aligned images. The SupCon loss helps bring the samples of bias aligned and bias conflicting groups within the same class together in the feature space, thus facilitating improved learning of their representations, as the resultant bias-conflicting images may deviate from the original data distribution as well as those of the generated bias-aligned samples.

% Please add the following required packages to your document preamble:
% \usepackage{booktabs}
\begin{table}[]
\centering
\caption{\textbf{Role of the SupCon loss on the performance of Clustered Dreambooth in case of Waterbirds}, for both the original and severely biased variants. Recall that $\beta=1$ denotes only using the CE loss, while $\beta=0.5$ refers to both losses having equal weight. While the performance on the original dataset does not show any significant change, the supcon loss improves accuracies for the high bias ratio version of the dataset.}
\label{tab:wb-supcon}
\begin{tabular}{@{}ll|ll|ll@{}}
\toprule
\multicolumn{2}{c|}{$\beta$} & \multicolumn{2}{c|}{\begin{tabular}[c]{@{}c@{}}Waterbirds\\ Original\end{tabular}} & \multicolumn{2}{l}{\begin{tabular}[c]{@{}l@{}}Waterbirds\\ Bias Ratio=0.999\end{tabular}} \\ \midrule
\multicolumn{1}{l|}{Stage 1} & Stage 2 & \multicolumn{1}{l|}{WGA} & AGA & \multicolumn{1}{l|}{WGA} & AGA \\ \midrule
\multicolumn{1}{l|}{$1$} & $1$ & \multicolumn{1}{l|}{$88.2$} & $90.2$ & \multicolumn{1}{l|}{$83.3$} & $88.0$ \\
\multicolumn{1}{l|}{$1$} & $0.5$ & \multicolumn{1}{l|}{$88.1$} & $90.2$ & \multicolumn{1}{l|}{$83.8$} & $88.0$ \\
\multicolumn{1}{l|}{$0.5$} & $1$ & \multicolumn{1}{l|}{$88.2$} & $90.2$ & \multicolumn{1}{l|}{$84.0$} & $88.0$ \\
\multicolumn{1}{l|}{$0.5$} & $0.5$ & \multicolumn{1}{l|}{$88.1$} & $90.2$ & \multicolumn{1}{l|}{$84.2$} & $88.5$ \\ \bottomrule
\end{tabular}
\end{table}

% Please add the following required packages to your document preamble:
% \usepackage{booktabs}
\begin{table}[]
\centering
\caption{\textbf{Role of the SupCon loss on the performance of Clustered Dreambooth in case of UTKFace}, for both the original and severely biased variants. Recall that $\beta=1$ denotes only using the CE loss, while $\beta=0.5$ refers to both losses having equal weight. While the performance on the original dataset shows slight increases, the supcon loss considerably improves accuracies for the high bias ratio version of the dataset.}
\label{tab:ut-supcon}
\begin{tabular}{@{}ll|ll|ll@{}}
\toprule
\multicolumn{2}{c|}{$\beta$} & \multicolumn{2}{c|}{\begin{tabular}[c]{@{}c@{}}UTKFace\\ Original\end{tabular}} & \multicolumn{2}{c}{\begin{tabular}[c]{@{}c@{}}UTKFace\\ Bias Ratio=0.999\end{tabular}} \\ \midrule
\multicolumn{1}{l|}{Stage 1} & Stage 2 & \multicolumn{1}{l|}{WGA} & AGA & \multicolumn{1}{l|}{WGA} & AGA \\ \midrule
\multicolumn{1}{l|}{$1$} & $1$ & \multicolumn{1}{l|}{$72.7$} & $88.2$ & \multicolumn{1}{l|}{$49.6$} & $80.7$ \\
\multicolumn{1}{l|}{$1$} & $0.5$ & \multicolumn{1}{l|}{$73.6$} & $83.4$ & \multicolumn{1}{l|}{$50.4$} & $80.8$ \\
\multicolumn{1}{l|}{$0.5$} & $1$ & \multicolumn{1}{l|}{$76.0$} & $83.5$ & \multicolumn{1}{l|}{$60.0$} & $80.4$ \\
\multicolumn{1}{l|}{$0.5$} & $0.5$ & \multicolumn{1}{l|}{$76.0$} & $83.5$ & \multicolumn{1}{l|}{$60.5$} & $80.8$ \\ \bottomrule
\end{tabular}
\end{table}

\noindent \textbf{Group-Balanced Finetuning}.
After pretraining on generated data, we finetune the classification layer on real data (LLR$_{\text{all}}$) (see Section~\ref{subsec: finetune}, main paper). To assess the benefits of group-balanced real data (sized to the smallest group), we explore:
a) \textit{Last Layer Retraining with Balanced Real Data} (LLR$_{\text{b}}$): Finetuning only the classification layer using the balanced real dataset instead of the full biased set.
b) \textit{Full Fine-Tuning with Balanced Real Data} (FT$_{\text{b}}$): Finetuning the entire network with the balanced real dataset.
Results show that FT$_{\text{b}}$ is beneficial only for CelebA, likely due to its larger size compared to Waterbirds and UTKFace (Table~\ref{tab:group-balance} for Clustered Dreambooth). In contrast, LLR$_{\text{b}}$ reduces worst group accuracy across all datasets, indicating that finetuning on the entire dataset yields better performance.

\begin{table}[]
\centering
\small
\caption{\textbf{Finetuning with group-balanced training data}. Upon manipulating the pretrained model with group-balanced training data, with and without finetuning the feature encoder $e^{\text{pre}}$ (FT$_{\text{b}}$ and LLR$_{\text{b}}$ respectively), we find that they are generally not advantageous, compared to LLR$_{\text{all}}$. The only exception is CelebA, where the WGA becomes $88.80\%$ for FT$_{\text{b}}$. CD: Clustered Dreambooth}
\label{tab:group-balance}
\begin{tabular}{@{}l|l|ll@{}}
\toprule
Dataset & Method & WGA & AGA \\ \midrule
\multirow{3}{*}{Waterbirds} & CD + LLR$_{\text{all}}$ & $87.8^{\pm 0.46}$ & $90.2^{\pm 0.14}$ \\
 & CD + LLR$_{\text{b}}$ & $87.4^{\pm 0.31}$ & $90.2^{\pm 0.12}$ \\
 & CD + FT$_{\text{b}}$ & $87.7^{\pm 0.30}$ & $90.7^{\pm 0.01}$ \\ \midrule
\multirow{3}{*}{CelebA} & CD + LLR$_{\text{all}}$ & $85.0^{\pm 0.35}$ & $88.6^{\pm 0.28}$ \\
 & CD + LLR$_{\text{b}}$ & $83.9^{\pm 1.36}$ & $88.2^{\pm 0.28}$ \\
 & CD + FT$_{\text{b}}$ & $88.8^{\pm 1.32}$ & $91.4^{\pm 0.39}$ \\ \midrule
\multirow{3}{*}{UTKFace} & CD + LLR$_{\text{b}}$ & $76.0^{\pm 1.22}$ & $83.5^{\pm 0.35}$\\
 & CD + LLR$_{\text{b}}$ & $70.20^{\pm 1.47}$ &  $83.78^{\pm 1.66}$\\
 & CD + FT$_{\text{b}}$ & $74.6^{\pm 2.55}$ & $83.8^{\pm 0.20}$ \\ \bottomrule
\end{tabular}
% \vspace{-0.5cm}
\end{table}

% Please add the following required packages to your document preamble:
% \usepackage{multirow}

\noindent \textbf{Effect of Scale.}
We generate 5,000 images per group for each dataset using a generative model to pretrain the classification network, selecting the top $75\%$ based on CLIP-based filtering (eq.~\ref{eq:2}, main paper). Table~\ref{tab:scale} presents the effect of total images generated on classification performance for Waterbirds (Dreambooth and Clustered Dreambooth). Comparing results across 20K, 15K, 10K, and 5K samples, we observe no significant variation in accuracy with increasing scale. 

\begin{table}[]
    \centering
    \small
    \caption{\textbf{Effect of Scale} (i.e., number of images generated). We evaluate Waterbirds performance using Dreambooth and Clustered Dreambooth for 20K to 5K generated images (top $75\%$ retained). No significant performance drop is observed.}
    \label{tab:scale}
    % \resizebox{\linewidth}{!}{%
        \begin{tabular}{@{}l|ll|cc@{}}
        \toprule
        \multirow{2}{*}{Scale} & \multicolumn{2}{l|}{Dreambooth} & \multicolumn{2}{l}{Clustered Dreambooth} \\ \cmidrule(l){2-5} 
         & \multicolumn{1}{l|}{WGA} & AGA & \multicolumn{1}{l|}{WGA} & AGA \\ \midrule
        20K & \multicolumn{1}{l|}{$88.3$} & $90.3$ & \multicolumn{1}{l|}{$88.8$} & $90.4$ \\
        15K & \multicolumn{1}{l|}{$88.5$} & $90.5$ & \multicolumn{1}{l|}{$87.9$} & $90.4$ \\
        10K & \multicolumn{1}{l|}{$88.2$} & $89.9$ & \multicolumn{1}{l|}{$87.9$} & $90.25$ \\
         5K & \multicolumn{1}{l|}{$89.3$} & $90.1$ & \multicolumn{1}{l|}{$87.8$} & $90.2$ \\ 
        \bottomrule
        \end{tabular}%
    % }
\end{table}

\begin{figure}[ht]
    \centering
    \includegraphics[width=0.7\columnwidth, trim=10 20 900 20, clip]{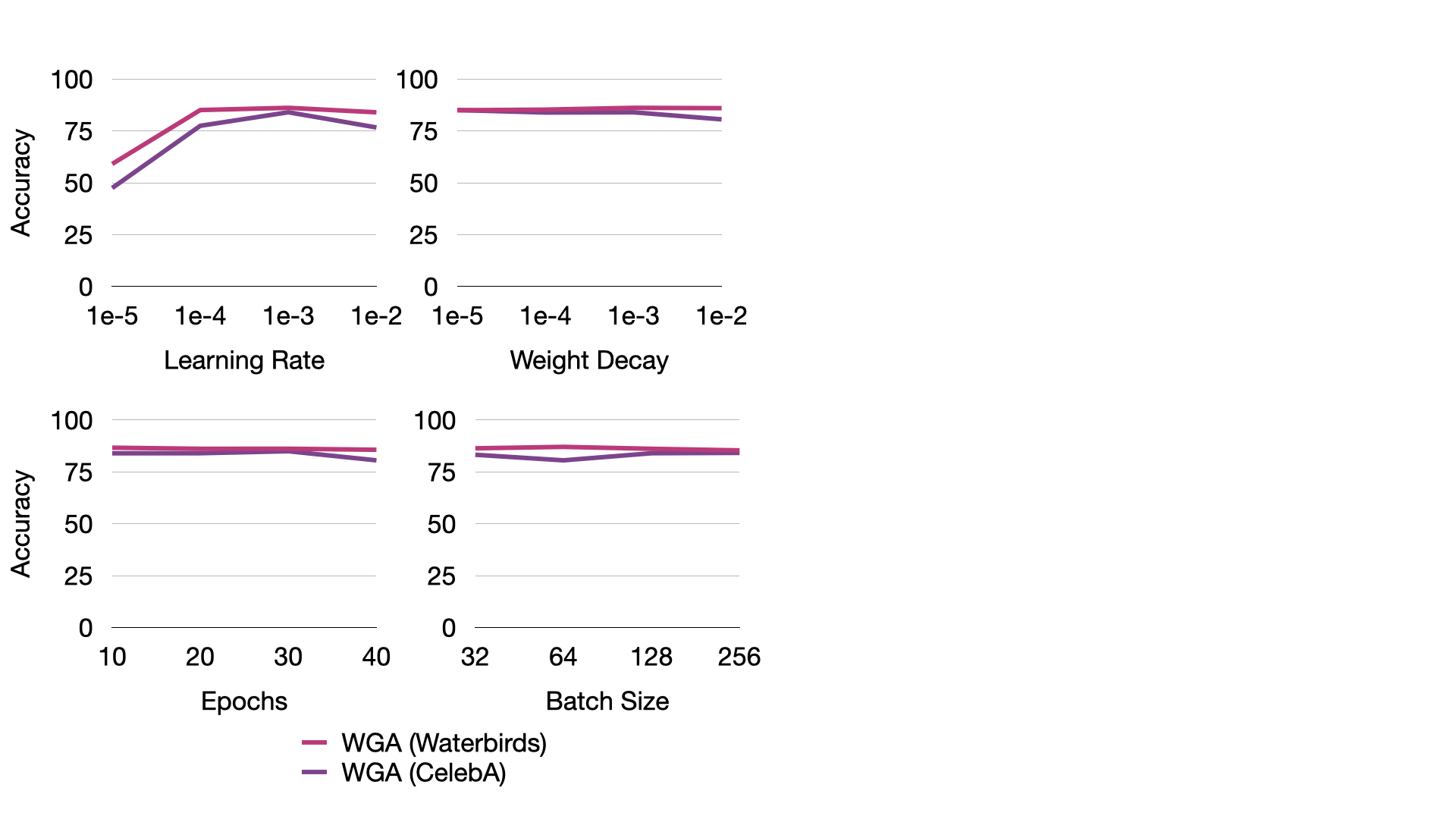}
    \caption{\textbf{Variations in worst group accuracies (WGA) with changing Hyperparameter values} for the pretraining stage on the CelebA~\cite{celeba} and Waterbirds~\cite{sagawa2019distributionally} datasets. Apart from a drastic fall at learning rate=$1e-5$, we do not observe any significant variation in the model performances across the observe range of hyperparameters.}
    \label{fig:hyp}
\end{figure}

\subsection{Hyperparameters for Classification}
\label{subsec:hyp}
While we do not experiment with hyperparameters during classification, using a learning rate of $1e-3$, weight decay of $1e-3$, batch size of $128$ and total epochs = $20$, we assess the variance of the classifier worst group performance on the pretraining stage of CelebA~\cite{celeba} and Waterbirds~\cite{sagawa2019distributionally} using the Clustered Dreambooth images. Specifically, we explore a set of learning rates : $\{1e-5, 1e-4, 1e-3, 1e-2\}$, a range of weight decay values: $\{1e-5, 1e-4, 1e-3, 1e-2\}$, training epochs: $\{10, 20, 30, 40\}$, and batch sizes: $\{32, 64, 128, 256\}$. We observe a drastic fall in accuracy scores with learning rate $= 1e-5$ for both datasets, which is expected, considering the low value of the learning rate. While the classifier performance on Waterbirds seems more consistent across hyperparameter values than that on CelebA, we do not observe any significant variation in model performance for different hyperparameters for any of the examined datasets. We plot our findings in Fig.~\ref{fig:hyp}. 

We next describe the hyperparameters used for synthetic data generation. For all the facial datasets, we include a negative prompt that discourages the model to generate grayscale images. While training the Dreambooth models, we use the default set of hyperparameters for all datasets. For LoRA-finetuning, we use the $r$ and $\alpha$ parameters of LoRA as $16$, and use no learning rate scheduler. The model is finetuned for 200 training steps. For all diffusion based methods, we set guidance-scale $=7.5$ and the number of timesteps to be $50$, except Clustered Dreambooth for UTKFace, where the number of timesteps is set to be $25$, based on manual inspection.  NVIDIA RTX A5000 card was used to generate synthetic images and train the classifier model.

\subsection{Quality of Representations learnt via Synthetic Data.} 
\label{subsec: fid}
To assess the similarities between the real and the generated distributions for each of the investigated approaches, we determine the Fr\'{e}chet Inception Distance (FID)~\cite{heusel2017gans} measured between the generated images and real images of each group of CelebA. Notably, Clustered Dreambooth images outperform all competing methods by a large margin, pointing to their similarity with the training images (see Table~\ref{tab:fid}). 

\begin{table*}[]
\centering
\small
\caption{\textbf{Evaluation of FID ($\downarrow$)}. The Clustered Dreambooth images outperform those of the other generative methods for all four groups of the CelebA dataset (Non-Blond Female (NF), Non-Blond Male (NM), Blond Female (BF), Blond Male (BM)).}
\label{tab:fid}
\begin{tabular}{@{}l|c|c|c|c@{}}
\toprule
Method & NF & NM & BF & BM \\ \midrule
Vanilla SD & $128.57$ & $133.53$ & $78.06$ & $92.48$ \\
FFR~\cite{qraitem2023fake} & $90.66$ & $98.16$ & $66.61$ & $100.27$ \\
% Textual Inversion & $96.57$ & $105.94$ & $89.52$ & $82.20$ \\
Dreambooth & $71.02$ & $69.18$ & $47.04$ & $58.51$ \\
Clustered Dreambooth & $\mathbf{58.72}$ & $\mathbf{56.48}$ & $\mathbf{38.50}$ & $\mathbf{46.90}$ \\ \bottomrule
\end{tabular}
\end{table*}

\subsection{Performance of Stage 1 pretraining on Test Data}
\label{subsec: stage1}
Recall that we pretrain the classification model on group-balanced synthetic images for each dataset. Evaluating on the test set, we find the worst group accuracy of such a model to be surprisingly high for the finetuned diffusion-based models, especially for Clustered Dreambooth, even without training the model on a single sample from the training set. This shows that the synthetic images resemble the real images closely. We present the results for the original training sets in Table~\ref{tab:stage1}.

\begin{table*}[]
\centering
\caption{\textbf{Classification Performance.} We report the classifier performance for Stage 1 (Generative Images Pretraining) and Stage 2 (Real Image Finetuning, denoted as LLR$_{\text{all}}$) for Vanilla SD, Dreambooth, and Clustered Dreambooth on the three datasets (original version). For Clustered Dreambooth, Stage 1 test accuracies are notably high across datasets.}
\label{tab:stage1}
\small
\begin{tabular}{c|c|ll|ll}
\toprule
\multirow{2}{*}{Dataset} & \multirow{2}{*}{Method} &  \multicolumn{2}{c|}{Stage 1} & \multicolumn{2}{c}{Stage 2} \\ \cmidrule{3-6} 
 &  & Worst & Average & Worst & Average \\ \midrule
\multirow{5}{*}{Waterbirds}
 & FFR$\dagger$~\cite{qraitem2023fake} &  $45.3$ & $71.3$ & $69.5$ & $84.0$ \\
 & Vanilla SD~\cite{ramesh2022hierarchical} & $65.4$ & $79.9$ & $74.6$ & $80.5$ \\
 & LoRA finetuning~\cite{gal2022image} &  $82.3$ & $89.3$ & $86.5$ & $89.9$ \\
 & Dreambooth~\cite{ruiz2022dreambooth} &  $85.7$ & $89.5$ & $89.3$ & $90.1$ \\
 & Clustered Dreambooth &  $87.4$ & $89.5$ & $88.1$ & $90.2$ \\ \midrule
\multirow{5}{*}{CelebA} 
 & FFR$\dagger$~\cite{qraitem2023fake} &  $48.9$ & $75.3$ & $68.9$ & $85.7$ \\
 & Vanilla SD~\cite{ramesh2022hierarchical} &  $78.3$ & $84.3$ & $76.40$ & $84.2$ \\
 & LoRA finetuning~\cite{gal2022image} &  $82.1$ & $87.1$ & $82.3$ & $87.2$ \\
 & Dreambooth~\cite{ruiz2022dreambooth} &  $81.7$ & $86.5$ & $82.1$ & $87.9$ \\
 & Clustered Dreambooth &  $83.9$ & $88.0$ & $84.1$ & $88.4$ \\
 \midrule
\multirow{5}{*}{UTKFace} 
 & FFR$\dagger$~\cite{qraitem2023fake} &  $66.1$ & $77.5$ & $67.4$ & $81.4$ \\
 & Vanilla SD~\cite{ramesh2022hierarchical} &  $68.0$ & $82.3$ & $62.0$ & $82.3$ \\
 & LoRA finetuning~\cite{gal2022image} &  $57.0$ & $82.0$ & $68.6$ & $85.6$ \\
 & Dreambooth~\cite{ruiz2022dreambooth} &  $56.2$ & $80.7$ & $57.9$ & $80.9$ \\
 & Clustered Dreambooth &  $66.9$ & $83.0$ & $76.0$ & $83.5$ \\
 \bottomrule
\end{tabular}
\end{table*}

\subsection{Qualitative Examples}
\label{subsec: qualitative2}
\begin{figure*}[ht]
    \centering
    \includegraphics[width=\textwidth, trim=0 10 0 0, clip]{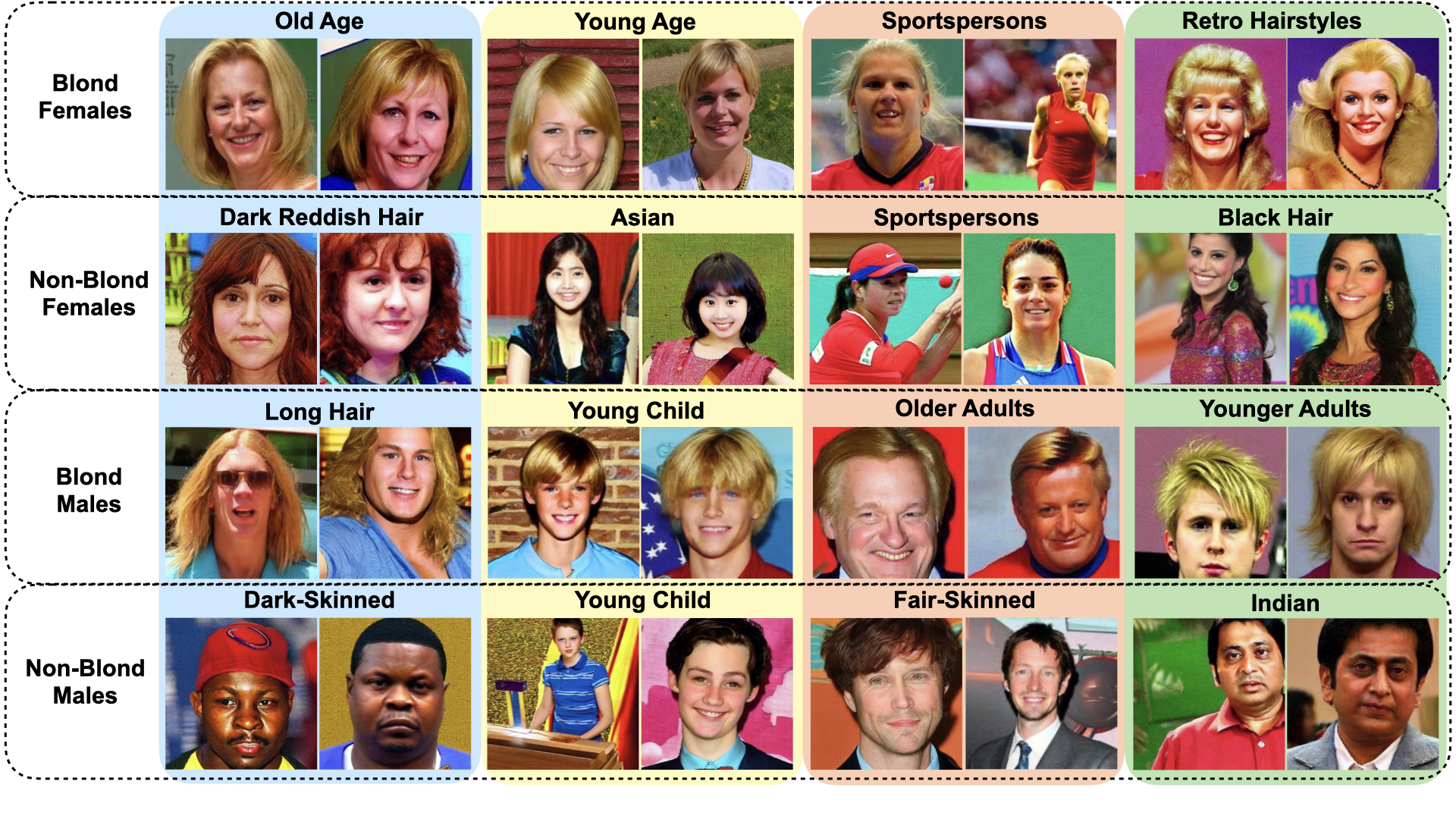}
    \caption{\textbf{Images generated from four different clusters of the CelebA groups.} We note the different characteristics followed by the images of a cluster for each group. We show two images per cluster and put the perceivable attributes seen in each cluster on the top (e.g., old age, young age, sportspersons and retro hairstyles for blond females).}
    \label{fig:cluster}
\end{figure*}

\noindent \textbf{Images generated by different models of Clustered Dreambooth}.
Here, we show examples from four different clusters in each group in the CelebA dataset~\cite{celeba}. We see that some of these models, while preserving the characteristics of its group (e.g. gender and hair color), captures different attributes like age, skin color, profession, hair color shades, etc. These variations can be clearly seen in Fig.~\ref{fig:cluster}, where the clusters are seen to generate older people, children, young adults, people of fair or dark skin, people from Indian descent, etc for the various groups. Interestingly, for all groups in CelebA, we also find clusters representing sportspeople as well, some of which are shown in the figure.

\begin{figure}[ht]
    \centering
    \includegraphics[width=0.5\columnwidth, trim=700 360 800 280, clip]{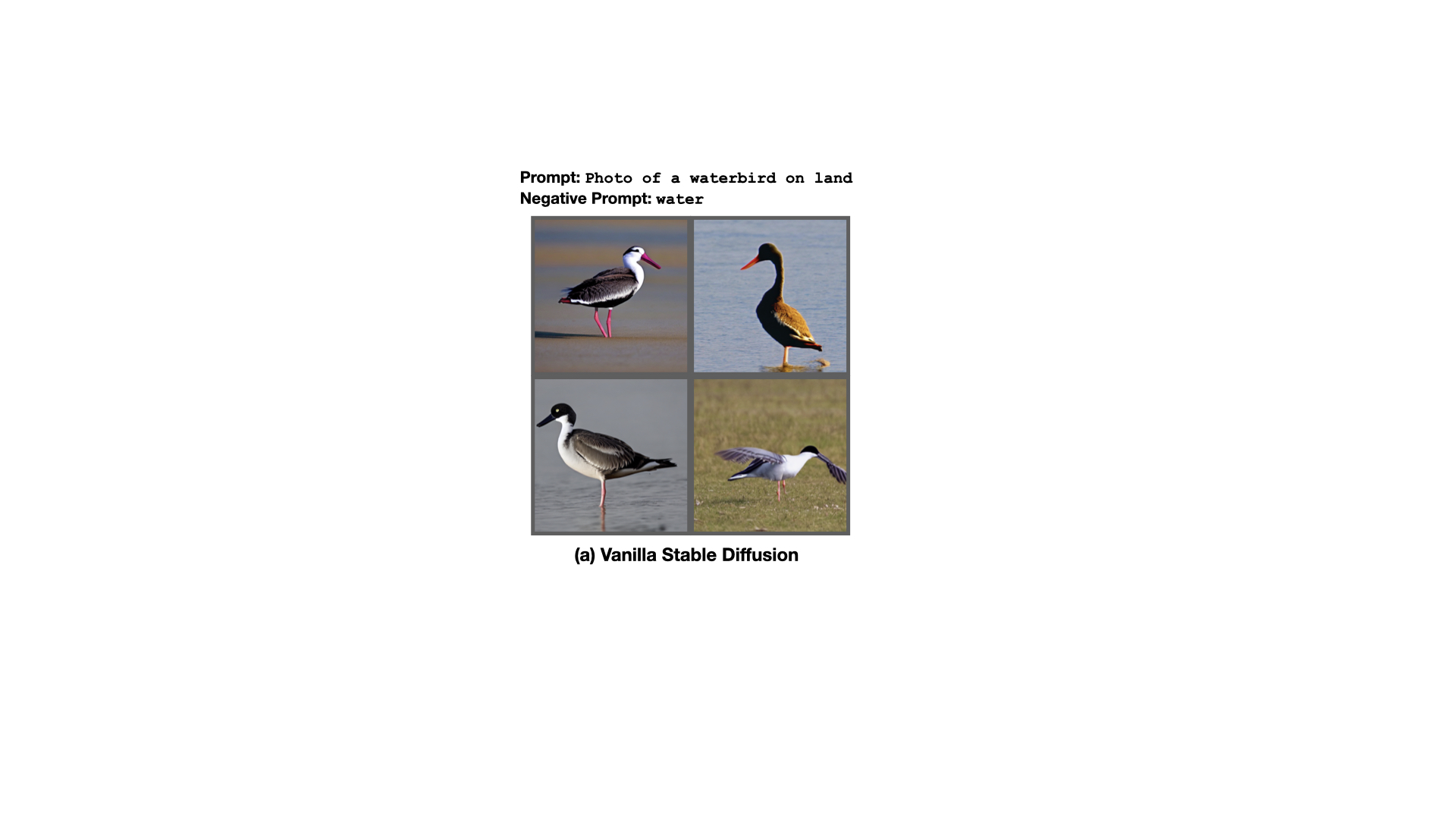}
    \caption{\textbf{Waterbird on Land Images generated the vanilla Stable Diffusion 1.4.} We note that the model often fails to follow the prompt instruction, and many images contain water even when explicitly prohibiting it in the negative prompt.}
    \label{fig:wb}
\end{figure}

\noindent \textbf{Vanilla SD failures in Waterbirds}
Vanilla SD models often fail to follow prompts accurately, perhaps because of inherent biases embedded in them. For example, when the model is instructed to generate `\texttt{Photo of a waterbird on land}', many of the generated images have water in them, even when we set a negative prompt `\texttt{water}'. This problem is highlighted in Fig.~\ref{fig:wb}, where more than $50\%$ of the images have water in the images. Moreover, while we agree that the generations are aesthically of high quality, the Waterbirds~\cite{sagawa2019distributionally} dataset itself is created by pasting bird images from the Caltech-UCSD Birds-200-2011 (CUB) dataset~\cite{welinder2010caltech} into images from the Place dataset~\cite{zhou2017places}, which often gives the images from the original dataset an unnatural look. Thus, this leads to a domain mismatch between the training and generated images, explaining the lower worst group accuracies of Vanilla SD.

\begin{figure}[ht]
    \centering
    \includegraphics[width=0.50\columnwidth, trim=700 380 790 280, clip]{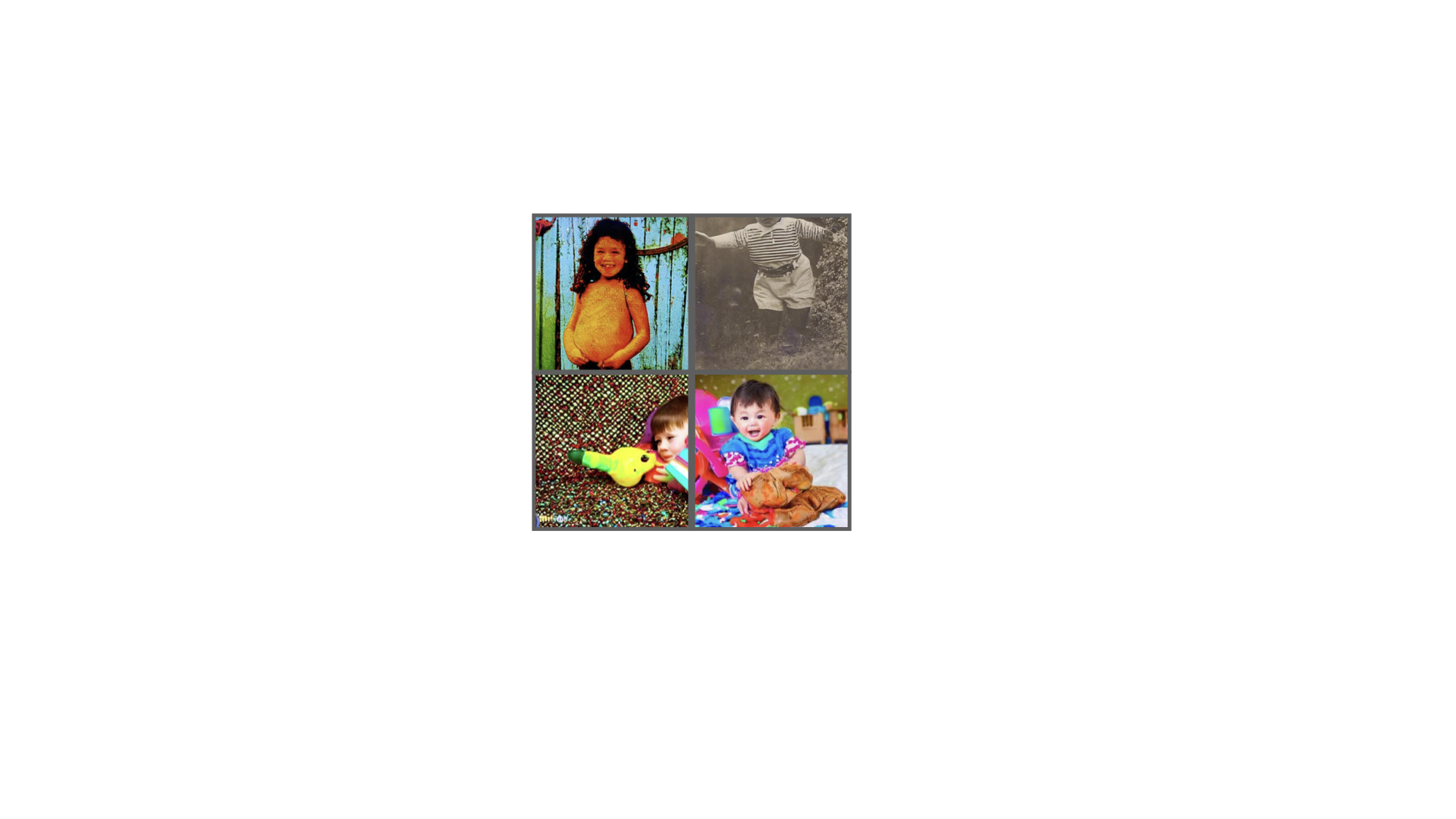}
    \caption{\textbf{Female Children images generated by Clustered Dreambooth for the high bias variant of UTKFace}. We note that the model often generates grayscale images, out-of-domain images, and also irrelevant ones.}
    \label{fig:utkfailure}
\end{figure}

\noindent \textbf{Failure cases of Clustered Dreambooth for UTKFace (Bias Ratio $\mathbf{0.999}$)}. As observed in Table~\ref{tab:main-table} (main paper), Clustered Dreambooth's performance suffers for the severely biased version of UTKFace compared to other methods. The worst group is Female Children, with an accuracy of $60.5 \%$, which is $15.5\%$ lower than that of the original data variant. We examine the reasons behind this performance drop by manually inspecting the Female Children images, and find that while many images are grayscale (inspite of having the word grayscale in the negative prompt), some are of Male Children. Alarmingly, some images fail to follow the training data domain, even though they correctly belong to Female Children. Such images get selected by our CLIP-Score as they accurately reflect the group description, and in the absence of the training images for the given group, we cannot compute the CLIP-Centroid Score to ensure that the selected images are as close to the training data domain as possible. We believe such issues potentially contribute towards the low worst group accuracies for the UTKFace Female Children group. Example images are shown in Fig.~\ref{fig:utkfailure}.

\begin{figure*}[ht]
    \centering
    \includegraphics[width=\textwidth, trim=0 0 0 0, clip]{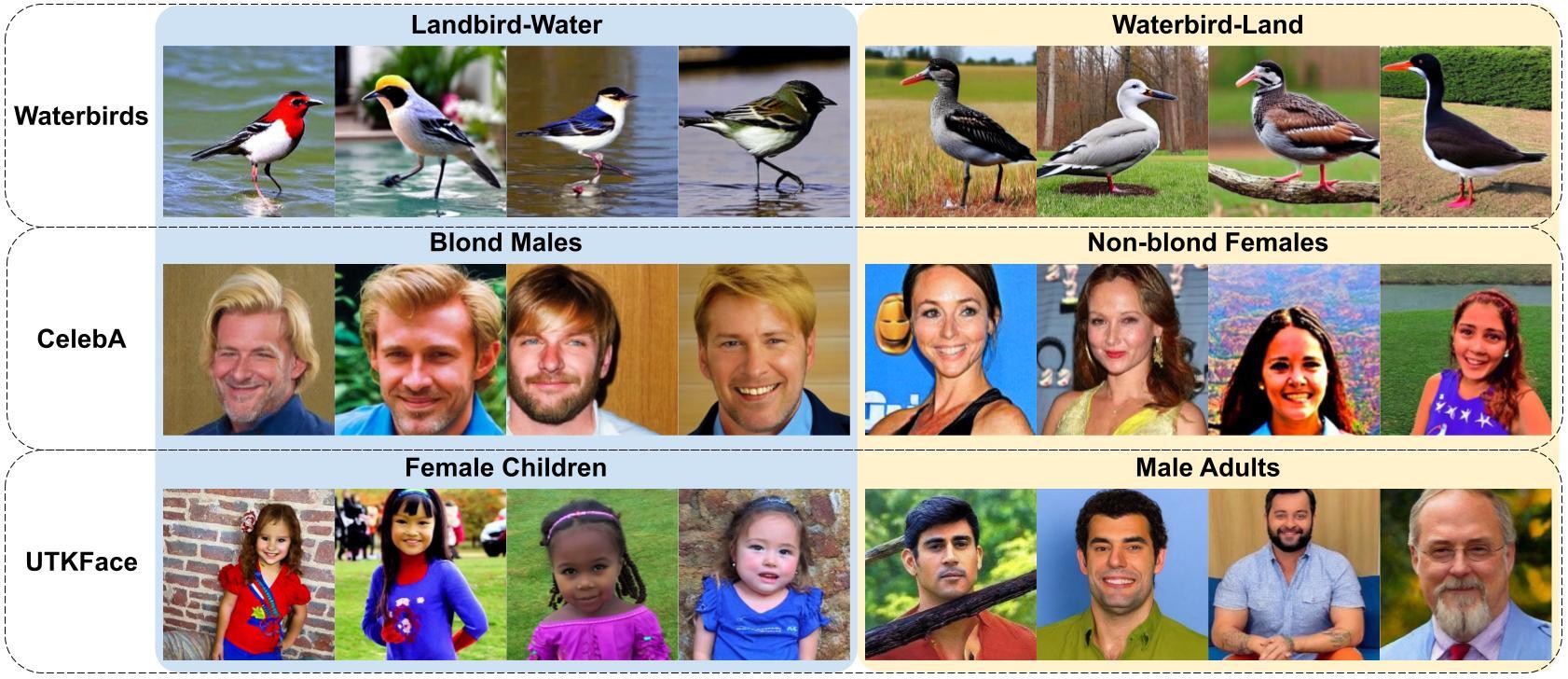}
    \caption{\textbf{Bias Conflicting samples generated for Bias Ratio$\mathbf{=0.999}$} using Clustered Dreambooth. Here we note that even in the absence of finetuned generative models for the bias-conflicting groups, the images generated from the models finetuned on the bias-aligned groups closely tend to follow the distribution of the dataset, while maintaining the requirements of the target group.}
    \label{fig:999}
\end{figure*}

\noindent \textbf{Generated Bias-Conflicting Samples from Clustered Dreambooth (Bias Ratio $\mathbf{=0.999}$)}. Recall that for bias ratio $0.999$, bias-conflicting images are generated using models trained on bias-aligned groups (e.g., Blond Males are generated from the model trained on Blond Females) for LoRA-finetuning, Dreambooth and Clustered Dreambooth. We present samples generated by the Clustered Dreambooth from the bias-conflicting groups of each dataset in Fig.~\ref{fig:999}, and observe that the generated images resemble the distribution of the input dataset, while reflecting the requirements of the target group.

\begin{figure*}[ht]
    \centering
    \includegraphics[width=\textwidth, trim=0 0 0 0, clip]{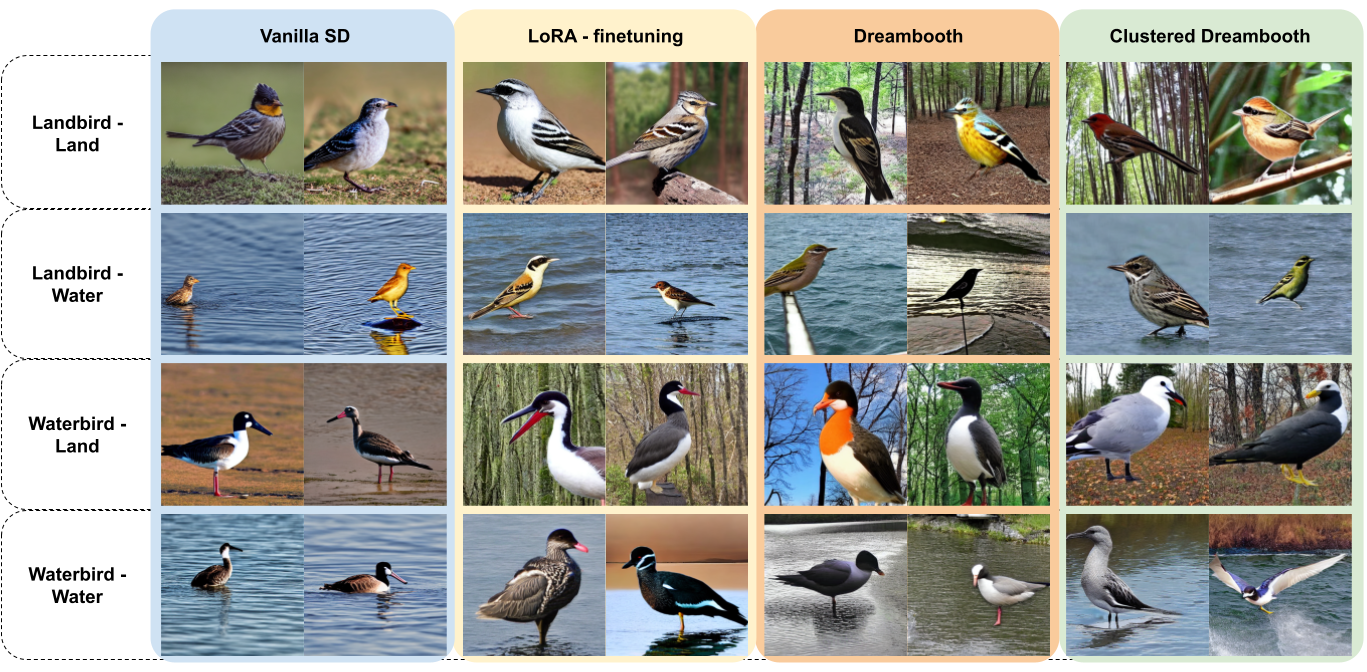}
    \caption{\textbf{Group-wise images for Waterbirds} generated by Vanilla SD, LoRA-finetuned SD, Dreambooth and Clustered Dreambooth.}
    \label{fig:waterbird}
\end{figure*}

\begin{figure*}[ht]
    \centering
    \includegraphics[width=\textwidth, trim=0 0 0 0, clip]{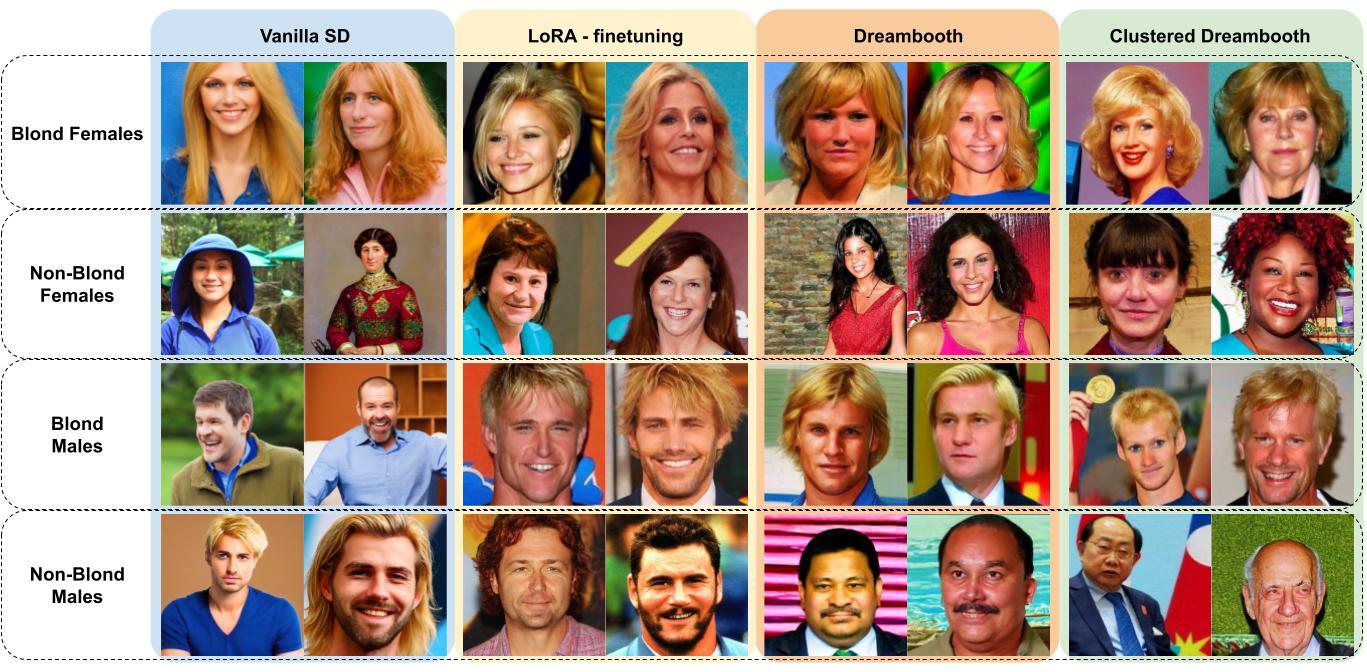}
    \caption{\textbf{Group-wise images for CelebA} generated by Vanilla SD, LoRA-finetuned SD, Dreambooth and Clustered Dreambooth.}
    \label{fig:celeba}
\end{figure*}

\begin{figure*}[ht]
    \centering
    \includegraphics[width=\textwidth, trim=0 0 0 0, clip]{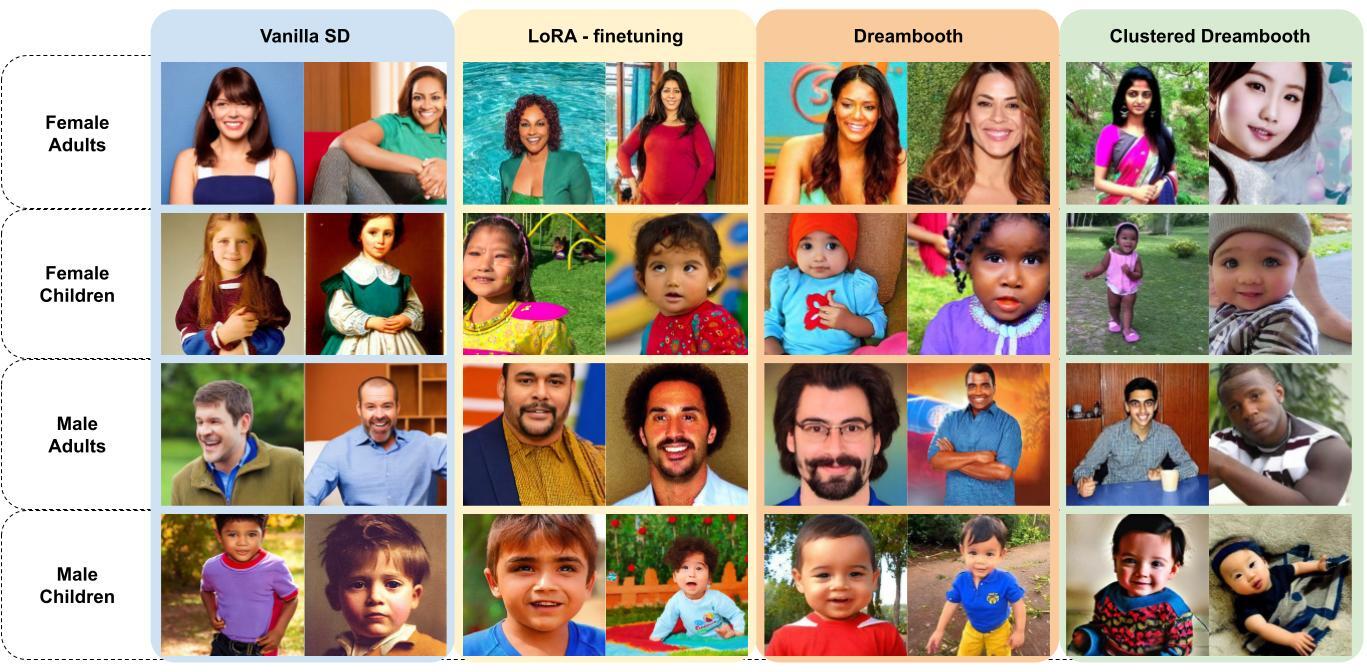}
    \caption{\textbf{Group-wise images for UTKFace} generated by Vanilla SD, LoRA-finetuned SD, Dreambooth and Clustered Dreambooth.}
    \label{fig:utkface-1}
\end{figure*}

\noindent \textbf{Images generated by each Diffusion-based Mechanism}. Here, we show the images generated by a) Vanilla SD, b) LoRA-finetuned SD, c) Dreambooth, and d) Clustered Dreambooth for each of the datasets Waterbirds, CelebA and UTKFace (all original versions), for each of the groups present in them. Figures ~\ref{fig:waterbird}, \ref{fig:celeba} and \ref{fig:utkface-1} present the examples for the three datasets (Waterbirds, CelebA and UTKFace) respectively.

\newpage
\bibliography{main.bib}

\end{document}